\documentclass{article}

\usepackage{microtype}
\usepackage{graphicx}
\usepackage{makecell}
\usepackage{xspace}
\usepackage{xcolor}
\usepackage{wrapfig}
\usepackage{subcaption}
\usepackage{booktabs} 

\usepackage{hyperref}

\usepackage[accepted]{icml2026}

\usepackage{amsmath}
\usepackage{multirow}
\usepackage{amssymb}
\usepackage{mathtools}
\usepackage{amsthm}
\usepackage{hyperref}

\usepackage{fontawesome5}
\def\changes{\textcolor{black}}

\usepackage[capitalize,noabbrev]{cleveref}

\theoremstyle{plain}

\theoremstyle{definition}

\theoremstyle{remark}

\usepackage[textsize=tiny]{todonotes}

\icmltitlerunning{Model-Based Diffusion Sampling for Predictive Control in Offline Decision Making}

\begin{document}

\twocolumn[
  \icmltitle{Model-Based Diffusion Sampling for\\ Predictive Control in Offline Decision Making}



  \icmlsetsymbol{equal}{*}
  \icmlsetsymbol{equaladvising}{$\dagger$}

  \begin{icmlauthorlist}
    \icmlauthor{Haldun Balim}{aff1}
    \icmlauthor{Na Li}{equaladvising,aff1}
    \icmlauthor{Yilun Du}{equaladvising,aff1}
  \end{icmlauthorlist}

  \icmlaffiliation{aff1}{Harvard University}

  \icmlcorrespondingauthor{Haldun Balim}{hbalim@fas.harvard.edu}
    \vspace{-0.1cm}
      \begin{center}
        \href{https://haldunbalim.github.io/MPDiffuser}{\faGlobe\ Website}
        \hspace{0.5cm}
    \href{https://github.com/haldunbalim/MPDiffuser}{\faGithub\ Code}
      \end{center}

  \icmlkeywords{Machine Learning, ICML}

  \vskip 0.3in
]



\printAffiliationsAndNotice{$^\dagger$Equal advising}

\begin{abstract}
Offline decision-making via diffusion models often produces trajectories that are misaligned with system dynamics, limiting their reliability for control. We propose \emph{Model Predictive Diffuser} (MPDiffuser), a compositional diffusion framework that combines a diffusion planner with a dynamics diffusion model to generate task-aligned and dynamically plausible trajectories. MPDiffuser interleaves planner and dynamics updates during sampling, progressively correcting feasibility while preserving task intent. A lightweight ranking module then selects trajectories that best satisfy task objectives. The compositional design improves sample efficiency and adaptability by enabling the dynamics model to leverage diverse and previously unseen data independently of the planner. Empirically, we demonstrate consistent improvements over prior diffusion-based methods on unconstrained (D4RL) and constrained (DSRL) benchmarks, and validate practicality through deployment on a real quadrupedal robot.
\end{abstract}

\vspace{-20pt}
\section{Introduction}
A central challenge in decision-making is designing policies that are both effective and reliable. Classical approaches address this through optimization, but are often limited by modeling assumptions and computational complexity~\citep{rawlings2017model}. In contrast, recent work has shown that data-driven generative models can achieve the same goal by sampling complex behaviors from available data~\citep{chi2023diffusion, janner2022planning, wang2024planning, pearce2023imitating, wang2022diffusion, chen2021decision, ajay2023conditional}. These methods are particularly appealing when interaction is costly or unavailable~\citep{janner2022planning}.

Offline decision-making formalizes this setting, requiring policies to be learned solely from collected data without further interaction~\citep{Figueiredo_Prudencio_2024}. While generative models provide a flexible framework for synthesizing candidate trajectories in this regime, existing approaches face important limitations. In particular, they struggle to effectively leverage suboptimal or heterogeneous data, often reproducing undesirable behaviors present in the dataset~\citep{hester2018deep, cheng2018fast}. Moreover, without explicit mechanisms to enforce dynamics consistency, these methods offer limited reliability and weak safety guarantees during execution~\citep{garcia2015comprehensive}.


In many real-world domains, including robotics~\citep{amodei2016concrete}, healthcare~\citep{yu2021reinforcement}, and autonomous driving~\citep{schwarting2018planning}, policies must satisfy safety constraints in addition to achieving task goals~\citep{dulac2021challenges, garcia2015comprehensive}. Enforcing safety offline is particularly challenging, as constraints must be satisfied without further interaction. Classical safe RL methods~\citep{achiam2017constrained, tessler2018reward, fujimoto2019off, kumar2020conservative} often fail under distribution shift, leading to conservative or unsafe behavior. In contrast, classical control methods enforce safety through short-horizon planning with explicit constraints~\citep{bemporad2007robust, rawlings2017model}. In this spirit, diffusion-based trajectory generation~\citep{janner2022planning, ajay2023conditional} provides a natural mechanism for producing diverse candidate rollouts, but existing methods operate directly in data space without enforcing system dynamics, limiting the reliability of the sampled trajectories.

\begin{figure*}[t]
    \centering
    \includegraphics[width=0.95\linewidth]{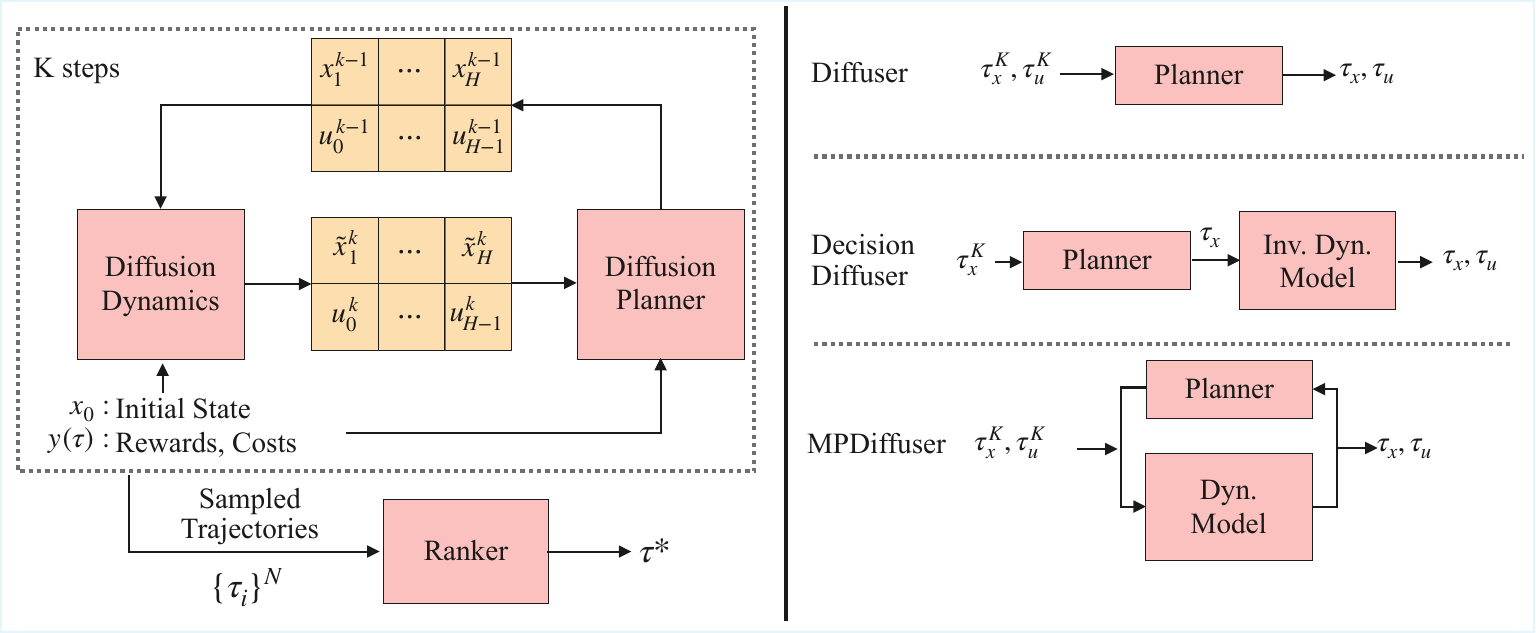}
    \caption{\textbf{Framework Overview.}
\textit{Left:} MPDiffuser, which couples a diffusion planner with a diffusion dynamics model, complemented by a ranking module.
\textit{Right:} Comparison highlighting key differences between our method and prior trajectory-level diffusion methods.}
    \label{fig:framework}
\vspace{-12pt}
\end{figure*}

\textbf{Contributions.} Motivated by these challenges, we propose \textit{Model Predictive Diffuser (MPDiffuser)}, a model-based compositional framework for offline decision making that combines three components: (i) a diffusion \emph{planner} that generates diverse, task-aligned trajectories; (ii) a diffusion \emph{dynamics model} that refines states to enforce consistency with system dynamics; and (iii) a \emph{ranker} that selects trajectories satisfying task-specific objectives and constraints.

MPDiffuser employs an alternating sampling scheme in which task-aligned proposals are repeatedly corrected by a diffusion dynamics model during sampling. This design balances task fidelity with dynamics feasibility and admits a theoretical interpretation as approximating a distribution combining planner priors with dynamics consistency. In contrast to prior diffusion methods that rely on inverse dynamics models~\citep{ajay2023conditional} or post-hoc forward models~\citep{zhou2025diffusion}, MPDiffuser directly models both states and actions, with the dynamics model acting as an active, dynamics aligned component of the sampling process. 

Empirically, MPDiffuser achieves consistent improvements in feasibility, safety, and decision quality across both unconstrained (D4RL) and constrained (DSRL) benchmarks. The compositional design improves sample efficiency by allowing the dynamics model to exploit low-quality and heterogeneous data, enables rapid adaptation to changes in system dynamics, and supports flexible integration of objectives and constraints via trajectory ranking. We further demonstrate scalability to visual domains and validate practical deployment on a quadrupedal robot.

\vspace{-10pt}
\section{Background \& Problem Setup}
\vspace{-5pt}

\subsection{Problem Setup}\label{sec:setup}

We consider a finite-horizon constrained Markov decision process (CMDP) defined by the tuple $(\mathcal{X},\mathcal{U},P,r,c,b,T)$, where $\mathcal{X}$ is the state space, $\mathcal{U}$ the action space, $P(x’ \mid x,u)$ the transition kernel, $r:\mathcal{X}\times\mathcal{U}\to\mathbb{R}$ the reward function, $c:\mathcal{X}\times\mathcal{U}\to\mathbb{R}+^m$ a vector of costs, $b \in \mathbb{R}+^m$ the cost budget, and $T$ the horizon. The objective is to find a policy $\pi$ that maximizes expected cumulative reward while satisfying the cost constraints:

\vspace{-20pt}
\begin{align*}
\max_\pi \mathbb{E}\!\left[\sum_{t=0}^{T-1} r(x_t,u_t)\right] \text{ s.t. }\mathbb{E}\!\left[\sum_{t=0}^{T-1} c_j(x_t,u_t)\right]&\le b_j,\\ j&\in\mathbb{I}_{[1,m]}.
\end{align*}

We consider the \emph{offline} setting, where interaction with the environment is not available and learning proceeds from a fixed dataset $\mathcal{D} = \{\xi_i\}_{i=1}^N$ of trajectories, with $\xi_i = \{(x_t, u_t, r(x_t,u_t), c(x_t,u_t))\}_{t=0}^T$. The available trajectories may be collected from multiple policies and can be suboptimal or unsafe. The objective is to learn a policy that maximizes cumulative reward while satisfying cost budgets.

A common approach extends value-based methods to jointly estimate reward and cost value functions and optimize a policy~\citep{lee2022coptidice}. With fixed data, however, value estimates deteriorate outside the dataset’s support: as the learned policy deviates from the behavior policies in the dataset, it induces poorly represented trajectories, leading to compounding errors and constraint violations.

An alternative viewpoint is to focus directly on synthesizing trajectories. Since rewards and costs depend on how actions drive the system’s evolution, full state–action rollouts provide a natural mechanism for evaluating task objectives and constraints. This trajectory-level perspective avoids unstable extrapolation, while offering a principled way to compute returns and costs. It therefore motivates generative approaches that explicitly model state–action trajectories.

\subsection{Trajectory Generation with Diffusion Models}

As introduced by \citet{sohl2015deep} and refined by \citet{ho2020denoising}, diffusion models are a class of generative models that approximate complex data distributions by reversing a gradual noising process. Due to their success in various domains, they have recently been applied to decision-making, where the objects of interest are state–action trajectories $\tau = (x_{1:H}, u_{0:H-1})$ of horizon $H$. By learning a diffusion model from the data, one can approximate the conditional distribution $p_\theta(\tau \mid x_0)$ and sample rollouts that resemble the dataset~\citep{janner2022planning}.  

The forward process incrementally perturbs a trajectory:
\begin{align*}
q_{k\mid k-1}(\tau^{k} \mid \tau^{k-1}) &= \mathcal{N}\!\left(\tau^{k}; \sqrt{1-\beta_k}\,\tau^k,\; \beta_k I \right),\\ q_{k\mid0}&=\mathcal{N}(\tau^k; \sqrt{\bar \alpha_k}, (1-\bar \alpha_k)I
\end{align*} 
where the variance schedule $\{\beta_k\}_{k=0}^K$ is fixed in advance and $\bar \alpha_k$ is defined by $\beta_k$. Accordingly, the reverse process seeks to undo this corruption using the score function:
\begin{align*}
s_{\theta}(\tau^k,k) &\;\approx\; \nabla_{\tau^k} \log q_k(\tau^{k}),\\
q_k(\tau^{k}) &= \int q_{k\mid 0}(\tau^{k}\mid\tau^{0})\,p_{\text{data}}(\tau^{0})\,d\tau^{0}.
\end{align*}
Intuitively, this score describes how likely a noisy sample $\tau^k$ is under the data distribution. Accordingly, the corresponding reverse transition is then given by:

\begin{align} 
\scalebox{0.93}{$
p_{k-1\mid k}(\tau^{k-1} \mid \tau^k) = \mathcal{N}\left(\tau^{k-1}; \frac{\tau^k }{\sqrt{\alpha_k}}  + \frac{\beta_k }{\sqrt{\alpha_k}}\, s_\theta(\tau^k,k) , \sigma_k^2 I\right), \notag 
$}
\end{align} 

where $\alpha_k$ and $\sigma_k$ are functions of $\beta_k$. Starting from Gaussian noise, clean trajectories are recovered by sampling from this reverse distribution. In practice, the score function is not estimated directly but learned implicitly through a \emph{noise prediction} objective. A trajectory $\tau$ is corrupted into $\tau^k$ by adding Gaussian noise $\epsilon \sim \mathcal{N}(0,I)$. A neural network $\epsilon_\theta$ is trained to recover this injected noise:
\[
\mathcal{L}(\theta) = \mathbb{E}_{\tau,k,\epsilon}\big[\;\|\epsilon - \epsilon_\theta(\tau^k,k)\|^2\;\big],
\]
which reduces to score matching under a simple reparameterization, ensuring that $\epsilon_\theta$ learns the score function.

\paragraph{Conditional generation.}  
In many applications, reproducing typical trajectories is not sufficient: we often require rollouts that are task-aligned. This motivates \emph{conditional trajectory generation}, where the model learns $p_\theta(\tau \mid x_0,y)$ for some condition $y(\tau)$, such as a target return or cost budget. A practical mechanism for enforcing such conditions is \emph{classifier-free guidance}~\citep{ho2022classifier}, which combines unconditional and conditional noise predictors:

\begin{align}\label{eq:cfg}
\scalebox{0.95}{$
\hat{\epsilon} = \epsilon_\theta(\tau^k,\emptyset,k)
+ \omega \Big(\epsilon_\theta(\tau^k,y(\tau),k) - \epsilon_\theta(\tau^k,\emptyset,k)\Big),$}
\end{align}

where $\emptyset$ denotes a fixed null input token for the condition and $\omega>0$ controls the guidance strength. This yields trajectory samples aligned with task objectives while retaining coverage of the dataset. To train both pathways in a single model, one uses \emph{conditional dropout}: the condition $y(\tau)$ is randomly masked with probability $p$, controlled by a Bernoulli variable $\beta$:
\[
\mathcal{L}(\theta) = \mathbb{E}_{k,\;\tau,\;\epsilon,\;\beta\sim\mathrm{Bern}(p)}
\Big[\;\|\epsilon - \epsilon_\theta(\tau^k, \beta y(\tau), k)\|^2\;\Big].
\]
Here $\beta=0$ masks the condition and sets it to null token $\emptyset$. This allows training both a conditional and unconditional predictor simultaneously. At inference, the two are recombined via classifier-free guidance (CFG) as in eq.~\eqref{eq:cfg}.

\begin{figure}[t]
    \centering
    \includegraphics[width=\linewidth]{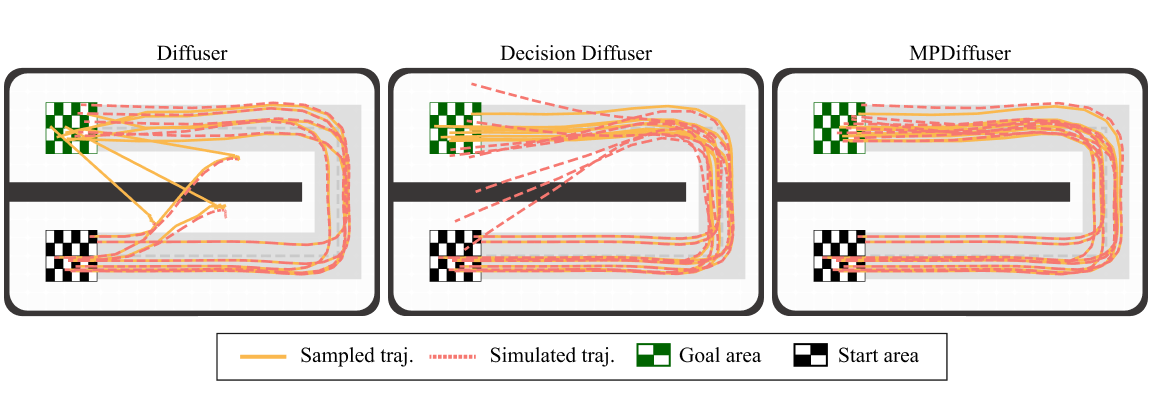}
    \caption{\textbf{Illustrative scenario:} We compare sampled state trajectories with open-loop simulations obtained by executing the sampled actions on a simple car model (cf. App.~\ref{sec:car}). Diffuser produces infeasible trajectories, and Decision Diffuser yields plausible states whose actions diverge when executed. In contrast, MPDiffuser generates trajectories that are faithful to system dynamics.}
    \label{fig:car}
\end{figure}

\section{Method}
Below, we introduce the components of our compositional framework, describe the sampling procedure, and show how it can be used for the constrained decision-making problem.

\subsection{Framework Components} 

We aim to generate trajectories that are high-reward, dynamically feasible, and constraint-compliant. A single model cannot balance these objectives (see Fig.~\ref{fig:car}): planners capture task intent but drift from dynamics, while dynamics models ensure feasibility but lack task guidance. To reconcile this, we introduce a compositional framework consisting of: a \emph{planner} that proposes task-aligned rollouts, a \emph{dynamics model} that enforces consistency with system transitions, and a \emph{ranker} that selects trajectories meeting objectives and safety. Each module is trained independently and combined only at inference, where their interaction yields trajectories aligned with both objectives and dynamics.

\paragraph{Planner Model.}
At the core of our framework, the planner acts as the trajectory generator—sampling diverse state–action sequences that pursue task objectives while capturing the variability encoded in the dataset. For this purpose, we train a conditional diffusion model over state-action trajectories $\tau_x=x_{1:H}, \tau_u=u_{0:H-1}$ given  initial state and conditioning vector $(x_0,\ y(\tau))$. We train a denoiser $\epsilon_\theta^{\mathrm{pl}}(\tau^k_x, \tau_u^k \mid k, x_0, y(\tau))$ via denoising score matching on sub-trajectories in $\mathcal{D}$ over a planning horizon $H$. The planner captures multi-modal intent, task structure, and dataset priors. The model is trained with the following loss:

\begin{align}
\scalebox{0.93}{$
\mathcal{L}^{\mathrm{pl}}(\theta)
=\mathbb{E}_{\tau, k, \epsilon,\beta}
\left[\;\big\|[\epsilon_x, \epsilon_u] - \epsilon_\theta^{\mathrm{pl}}\!\big(\tau^k_x, \tau_u^k\mid k, x_0, \beta y(\tau)\big)\big\|^2\;\right]. \notag $}
\end{align}

\paragraph{Dynamics Model.}
Complementing the planner, the dynamics model acts as the feasibility filter—refining state trajectories so that imagined rollouts remain faithful to the system’s underlying physics. In contrast to the planner, which models full state–action rollouts, the dynamics model is a conditional diffusion model over \emph{states only}, given the initial state $x_0$, an action sequence is corrupted by the same noise level as states $\tau_u^k$, and conditioning vector $y(\tau)$. By treating actions purely as inputs, the model dedicates its capacity to capturing state transitions, thereby yielding sharper dynamics consistency than the planner. We train our dynamics diffusion model $\epsilon_\vartheta^{\mathrm{dyn}}(\tau_x^k \mid \tau_u^k,k, x_0,y(\tau))$ using:
\begin{align}
\scalebox{0.93}{$
\mathcal{L}^{\mathrm{dyn}}(\vartheta)
=\mathbb{E}_{\tau, k,\epsilon,\beta}
\left[\;\big\|\epsilon_x - \epsilon_\vartheta^{\mathrm{dyn}}\!\big(\tau_x^k\mid \tau_u^k, k, x_0, \beta y(\tau)\big)\big\|^2\;\right].\notag $}
\end{align}
\paragraph{Ranker.}
The ranker is a practical task-aware module that evaluates sampled trajectories against desired criteria, selecting rollouts that achieve high reward, respect safety budgets. Within our framework, it is treated as a flexible scoring function $\rho(\tau)$ that assigns preference values to trajectories. This allows incorporation of both domain knowledge and data-driven objectives. Formally, given sampled trajectories $\{\tau_j\}$, the ranker outputs $\tau^\star = \arg\max_{\tau_j}\rho(\tau_j)$, with $\rho$ defined by the task e.g. maximizing return under constraints, or minimizing goal distance. This design balances flexibility and structure: when objectives are clearly specified, $\rho$ can be explicitly defined analytically, while in settings with implicit preferences, $\rho$ may be learned from data.

\subsection{Alternating Diffusion Sampling}
\label{sec:sampling-procedure}

To generate trajectories, we employ an alternating diffusion sampling scheme (Alg.~\ref{alg:alt-guided}) that decomposes denoising into two complementary updates: one enforcing dynamics feasibility and the other promoting task alignment. Starting from Gaussian noise, each reverse step first applies the dynamics model to refine states conditioned on the current actions, projecting them toward the manifold of feasible transitions, followed by the planner, which jointly denoises states and actions to restore task structure and dataset consistency.

Classifier-free guidance is applied in both steps, to encourage task conditioning. During the reverse process, the planner pushes samples toward task-aligned regions, while the dynamics model counteracts drift and enforces feasibility. The alternating composition thus functions like a dialogue: the planner expands trajectories toward task objectives, and the dynamics model regularizes them to system transitions. 

\begin{figure}[t]
\vspace{-10pt}
\begin{minipage}{\linewidth}
 \begin{algorithm}[H]
    \caption{Alternating Diffusion Sampling for Conditional Trajectory Generation}
    \label{alg:alt-guided}
    \begin{algorithmic}[1]
      \REQUIRE  Planner $\epsilon^{\mathrm{pl}}_\theta(\tau_x^k,\tau_u^k\mid k, x_0,y)$; Dynamics  $\epsilon^{\mathrm{dyn}}_\vartheta(\tau_x^k\mid\tau_u^k,k, x_0,y)$; guidance scale $\omega$; condition $y$; initial state $x_0$, temperature $\alpha$
      \STATE Initialize trajectory $\tau_x^k, \tau_u^k\sim\mathcal{N}(0,\alpha I)$
      \FOR{$k=K,\dots,1$}

    \item[] \(\triangleright\) \textcolor{lightgray}{Dynamics step: update states only}

        \STATE $\hat{\epsilon}_x \leftarrow
        \mathrm{CFG}_\omega\!\left(
        \epsilon_\vartheta^{\mathrm{dyn}}(\tau^k_x | \tau^k_u, k, x_0, y) 
        \right)$ \textcolor{gray}{(eq.~\eqref{eq:cfg})}
        \STATE $(\tilde \tau_x, \Sigma^{k-1}_x) \leftarrow \mathrm{Denoise}(\tau_x^k, \hat{\epsilon})$
      \item[] \(\triangleright\) \textcolor{lightgray}{Planner step: update states \& actions jointly}
        
        \STATE $\hat{\epsilon}_\tau \leftarrow
        \mathrm{CFG}_\omega\!\left(
        \epsilon_\theta^{\mathrm{pl}}(\tilde \tau_x, \tau_u^k\mid k,x_0, y) 
        \right)$  \textcolor{gray}{(eq.~\eqref{eq:cfg})}
        \STATE $(\mu^{k-1}_\tau, \Sigma^{k-1}_\tau) \leftarrow \mathrm{Denoise}(\tilde \tau_x, \tau_u^k, \hat{\epsilon}_\tau)$
        \STATE $\tau_x^{k-1}, \tau_u^{k-1}\sim \mathcal{N}(\mu^{k-1}_\tau, \alpha \Sigma^{k-1}_\tau)$
      \ENDFOR
      \STATE \textbf{return} $\tau=(\tau_x^0, \tau_u^0)$
    \end{algorithmic}
  \end{algorithm}

\end{minipage}
\vspace{-15pt}
\end{figure}

\subsection{Constrained Control with MPDiffuser}
Algorithm~\ref{alg:budget-sample} integrates trajectory sampling with budget-feasible selection: candidate rollouts from Algorithm~\ref{alg:alt-guided} are evaluated by reward and cost models, after which the ranker returns the highest-return feasible trajectory or the least-cost one if none satisfy the budget. As a modular component, the ranker can be tailored to diverse objectives—prioritizing rewards, enforcing constraints, or inducing task-specific skills. Notably, the use of return and cost scaling parameters enable adaptation without retraining, allowing MPDiffuser  to generate safer or more risk-tolerant behaviors as needed.
\begin{figure}[H]
\vspace{-15pt}
\begin{minipage}{\linewidth}
\begin{algorithm}[H]
  \caption{Cost Budget-Aware Trajectory Sampling}
  \label{alg:budget-sample}
  \begin{algorithmic}[1]
    \REQUIRE initial state $x_0$ , num. candidates $N$, condition $y$, remaining budget $b_{\mathrm{rem}}$, reward model $\hat r$, cost model $\hat c$, discount factor $\gamma$
    \STATE Sample $N$ trajectories $\{\tau^{(i)}\}_{i=1}^N \gets \textsc{Algo~\ref{alg:alt-guided}}(x_0,y)$
    \FOR{$i=1$ to $N$}
      \STATE $\hat J_i \gets \sum_{t=0}^{H-1}\gamma^t \hat r(x_t^{(i)},u_t^{(i)})$
      \STATE $\hat C_i \gets \sum_{t=0}^{H-1}\gamma^t \hat c(x_t^{(i)},u_t^{(i)})$
    \ENDFOR
    \STATE $\mathcal{F} \gets \{\,i \mid \hat C_i \le b_{\mathrm{rem}}\}$ \COMMENT{filter feasible trajectories}
    \IF{$\mathcal{F}\neq\emptyset$} 
        \STATE \textbf{return} highest return feasible trajectory
    \ELSE 
        \STATE \textbf{return} minimum cost trajectory
    \ENDIF
  \end{algorithmic}
\end{algorithm}
\end{minipage}
\vspace{-10pt}
\end{figure}

\subsection{Rationale behind the algorithm development}


Here, we provide a brief discussion on a theoretical rationale for our Alg.~\ref{alg:alt-guided}. For an extended discussion refer to Appendix~\ref{sec:theory}. We consider two distributions over trajectories. The former is the planner distribution
$p_{\mathrm{pl}}(\tau\mid x_0)$, induced by running a diffusion sampler with the learned planner model; this distribution captures task structure and preferences from demonstrations. The second is the dynamics distribution $p_{\mathrm{dyn}}(\tau\mid x_0)\propto\prod_{t} P(x_{t+1}\mid x_t,u_t)$, which assigns higher probability to trajectories consistent with the system transition kernel. To balance these two objectives, we seek a distribution $q$ close to the planner but with high dynamics likelihood. This constrained projection can be written as:
\[
\min_{q}\; \mathbb{E}_q[-\log p_{\mathrm{dyn}}(\tau\mid x_0)] \quad
\text{s.t.}\quad \mathrm{KL}(q\;\|\;p_{\mathrm{pl}})\le \varepsilon,
\]
Introducing a Lagrange multiplier $\lambda > 0$ for the KL constraint, we obtain the relaxed
objective:
\[
q^\ast(\tau\mid x_0)\;\propto\;p_{\mathrm{pl}}(\tau\mid x_0)\;p_{\mathrm{dyn}}(\tau\mid x_0)^\lambda.
\]
Directly characterizing $q^\ast$ is not possible, as it is an abstract construction combining
$p_{\mathrm{pl}}$ and $p_{\mathrm{dyn}}$, and we do not have samples from it to fit a diffusion
model. Nevertheless, sampling from $q^\ast$ can in principle be achieved via its score function
$s_{q^\ast}$, which determines the probability--flow dynamics. The exact score is intractable, but
by analogy with classifier guidance we approximate it as sum of scores:
\begin{equation}\label{eq:combined_score}
    s_{q^\ast}(\tau^k,k)\;\approx\;s_{p^{\mathrm{pl}}}(\tau^k,k)+\lambda\,s_{p^{\mathrm{dyn}}}(\tau^k,k),
\end{equation}
where $s_{q^\ast}$, $s_{p^{\mathrm{pl}}}$, $s_{p^{\mathrm{dyn}}}$ denotes the score function of the corresponding distributions. A natural approach is to sample using this combined score, but in practice such updates can be unstable because planner and dynamics gradients often differ in scale or curvature \changes{(cf. Sec.~\ref{sec:analysis})}. Our algorithm instead alternates between planner and dynamics updates. This design is
motivated by operator-splitting~\citep{hairer2006geometric,trotter1959product}, which approximate the flow of a combined system by alternating short steps under each component. Although both models share the same architecture and training data, the dynamics model focuses exclusively on state prediction, while the planner models both states and actions. This allows the dynamics model to capture transition structure more accurately, yielding a stronger state-consistency signal. Thus, alternating sampling combines the planner’s task alignment with the dynamics model’s precision, guiding sampling effectively toward $q^\ast$.

\vspace{-5pt}
\section{Experiments}
\vspace{-5pt}

We evaluate our method across diverse settings to demonstrate its effectiveness, versatility, and practicality. Our experimental evaluation includes: (1) \textbf{offline decision making} on D4RL benchmark tasks, including adaptation to novel dynamics, assessment of feasibility of the generated sequences, and leveraging random data for dynamics learning (Sec.~\ref{sec:exp-offline}); (2) \textbf{constrained offline decision making} on safety-critical DSRL benchmarks with cost constraints, together with a study on the Pendulum environment highlighting the importance of dynamic feasibility for ranking (Sec.~\ref{sec:exp-const-offline}); (3) a preliminary study extending our framework to handle visual inputs (Sec.~\ref{sec:vision}); (4) \textbf{real-world deployment} on a Unitree Go2 quadruped robot to demonstrate the practicality of MPDiffuser (Sec.~\ref{sec:exp-dog}); and (5) an \textbf{analysis} section examining the effect of alternating updates, comparison with combined-score updates, and limitations under cross-dataset distribution mismatch (Sec.~\ref{sec:analysis}).

Additional studies are deferred to the appendix, including:
(i) a \textbf{linear control system} for validation in a well-understood simple setting (App.~\ref{sec:linsys});
(ii) ablations on initial-state conditioning (App.~\ref{sec:condition}),
robustness to dynamics modeling errors (App.~\ref{sec:robust}),
conditioning in the dynamics model (App.~\ref{sec:dyn-cond}),
the impact of causal networks (App.~\ref{sec:causal}),
and sensitivity to guidance scale and number of samples (App.~\ref{sec:params}).

\subsection{Offline Decision Making}\label{sec:exp-offline}

\textbf{Results on Standardized Benchmarks (D4RL):} We evaluate MPDiffuser , in two configurations: (i) MPDiffuser  using a single trajectory sample, and (ii) MPDiffuser +Rank using multiple samples (64), where the ranker selects highest return trajectory among sampled candidates using a learned reward model. Experiments are conducted on the D4RL benchmark~\citep{d4rl}. We compare against standard baselines such as Behavior Cloning (BC) and Decision Transformer (DT)~\citep{chen2021decision}, \changes{a model-based offline RL algorithm (COMBO)~\citep{yu2021combo}}, as well as recent diffusion-based methods including IDQL~\citep{hansen2023idql}, Diffusion MPC (D-MPC)~\citep{zhou2025diffusion}, Decision Diffuser~\citep{ajay2023conditional} and Diffuser~\citep{janner2022planning}. To isolate the effect of the alternating sampling scheme, we include results using the planner alone.

Table~\ref{tab:d4rl} shows the average normalized returns on all considered tasks. The alternating planner--dynamics sampling yields trajectories that are better aligned with the dataset distribution, leading to improved performance even in unconstrained settings. The ranker adds a modest but consistent gain by selecting trajectories more closely matched to task objectives, with the effect most pronounced on domains that require longer-horizon planning such as Kitchen.

\begin{table*}[t]
\centering
\resizebox{\textwidth}{!}{
\begin{tabular}{ll|ccccccccccc}
\toprule
Dataset & Environment &  BC & DT & \changes{COMBO} &\changes{IDQL} & Diffuser & \makecell{Decision \\ Diffuser} & D-MPC & Planner & MPDiffuser  & MPDiffuser+Rank\\
\midrule
\multirow{3}{*}{\makecell{Med-Exp}} 
& Hopper &  $ 52.5 $  & $ 107.6 $  &\changes{111.1} & \changes{105.3} &  $ 107.2 $  & $ \mathbf{111.8}$ & &  $109.5$ & $ 109.9\pm 1.1 $ &  $110.4\pm 0.0$\\
& Walker2d &  $ 107.5 $  & $ 108.1 $  & \changes{103.3} & \changes{111.6}& $ 108.4 $  & $ 108.8 $  & &  $110.4$ & $ \mathbf{110.7}\pm 0.7 $  & $\mathbf{110.7 \pm 0.2}$  \\
& HalfCheetah  & $ 55.2 $  & $ 86.8 $  & \changes{$ 90.0$} & \changes{94.4} & $ 79.8 $  & $ 90.6 $ &  & $95.7$ & $ 96.9\pm 0.0 $ & $\mathbf{98.4} \pm 0.0$   \\
\midrule
\multirow{3}{*}{\makecell{Medium}} 
& Hopper  & $ 52.9 $  & $ 67.6 $& \changes{97.2} & \changes{63.1}  & $ 58.5 $  & $ 79.3 $ &$61.2$ & $97.6$ & $ 97.9\pm 0.3 $ & $\mathbf{98.4} \pm 0.4$   \\
& Walker2d  & $ 75.3 $  & $ 74.0 $  & \changes{$\mathbf{81.9}$} &\changes{80.2} & $ 79.7 $  & $ \mathbf{82.5}$ & $76.2$ & $75.9$ & $ 77.5 \pm 0.5$  & $77.6\pm 0.0$  \\
& HalfCheetah  & $ 42.6 $  & $ 42.6 $  & \changes{$\mathbf{54.2}$} & \changes{49.7} & $ 44.2 $  & $ 49.1$ & $46.0$& $ 47.6$ & $ 47.9\pm 1.6 $  & $47.9\pm 1.0$ \\
\midrule
\multirow{3}{*}{\makecell{Med-Replay}} 
& Hopper &  $ 18.1 $  & $ 82.7 $  & \changes{89.5}& \changes{82.4} & $ 96.8 $  & $ \mathbf{100.0}$ &$92.5$ & $92.1$ & $ 98.2 \pm 0.3 $ & $98.3\pm 0.7$  \\
& Walker2d  & $ 26.0 $  & $ 66.6 $  &\changes{56.0} & \changes{79.8} & $ 61.2 $  & $ 75.0 $  &$78.8$ & $71.8$ & $ \mathbf{81.5}  \pm 0.7$ & $81.2\pm0.8$  \\
& HalfCheetah  & $ 36.6 $  & $ 36.6 $  & \changes{$\mathbf{ 55.1} $} & \changes{45.1} & $ 42.2 $  & $ 39.3 $  &$41.1$ & $44.0$ & $43.4\pm 1.1 $ & $\mathbf{43.5}\pm0.5$   \\
\midrule
\multicolumn{2}{c|}{\textbf{Average}} & $ 51.9 $  & $ 74.7 $  &\changes{82.0}& \changes{79.1} &$ 75.3 $  & $ 81.8$ & & $ 82.7$ & $ 84.9 $ & $ \mathbf{85.1} $   \\
\midrule
Mixed & Kitchen & $ 51.5 $  & $ $  & & & $  $  & $ \mathbf{65} $ & $67.5$ & $57.2$ & $ 66.1\pm 1.7 $ & $ 66.9 \pm1 .7 $   \\
Partial & Kitchen & $ 38 $  & $ $  & & & $  $  & $ 57 $ & $73.3$ & $57.2$ & $ 67.9 \pm 2.6$ & $73.8 \pm 1.5$   \\
\midrule
\multicolumn{2}{c|}{\textbf{Average}} & $ 44.8 $  & & &  $ $  & $  $  & $ 61 $ &$\mathbf{70.4}$ & $57.2$ & $67.0 $ & $\mathbf{70.4}$   \\
\bottomrule
 \end{tabular}}
 \caption{\textbf{Performance on D4RL benchmark tasks.} We report normalized average scores with corresponding standard deviations under the standard D4RL evaluation protocol~\citep{d4rl}. Results are averaged over 5 independent runs, each evaluated on 50 rollouts. MPDiffuser outperforms prior baselines, while MPDiffuser+Rank provides further improvements by selecting higher-quality trajectories.}
 \label{tab:d4rl}
 \vspace{-18pt}
\end{table*}

\textbf{Leveraging Random Data for Dynamics Learning:} We consider \texttt{FetchPickAndPlace} task, where a robot arm must bring a block to a target location, with success defined as bringing block close enough to the goal position. Models are conditioned on the goal, and ranker picks trajectories by minimum block–goal distance. The planner is trained on $1000$ expert demonstrations, while the dynamics model uses additional trajectories obtained by applying random actions.

\begin{figure}[h]
  \begin{minipage}[t]{0.14\textwidth}
    \vspace{0pt}
    \centering
    \includegraphics[width=1\linewidth]{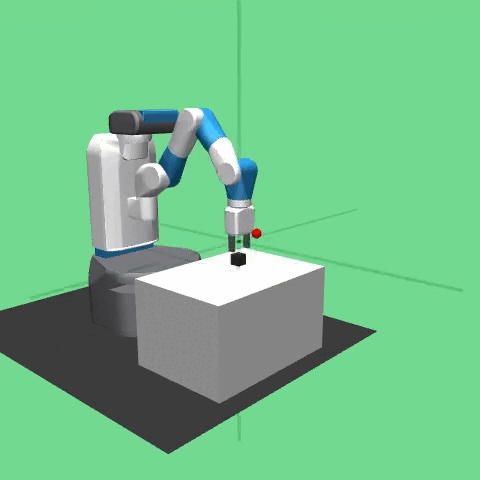}
    \captionsetup{type=figure}
    \captionof{figure}{\texttt{Fetch PickandPlace}}
  \end{minipage}
  \hfill
  \begin{minipage}[t]{0.31\textwidth}
    \centering
  \vspace{0pt}
    \captionsetup{type=table}
      \small
    \begin{tabular}{@{}ccc@{}}
      \toprule
      Num. rand. traj. & MPD & D-MPC \\
      \midrule
      0      & 0.75 & 0.60 \\
      2000   & 0.81 & 0.68 \\
      4000   & 0.79 & 0.56 \\
      6000   & 0.78 & 0.53 \\
      8000   & 0.82 & 0.55 \\
      10000  & 0.86 & 0.60 \\
      \bottomrule
    \end{tabular}
    \captionof{table}{\changes{\textbf{MPDiffuser can harness suboptimal data.} Success rate versus number of random trajectories for dynamics training.}}
    \label{tab:fetch}
  \end{minipage}
  \vspace{-15pt}
\end{figure}

Table~\ref{tab:fetch} reports success rates. Adding random trajectories for dynamics training improves performance, even though the planner relies solely on expert data. This demonstrates a key benefit of our compositional framework: while training planners require high-quality data, the dynamics model can effectively exploit inexpensive random data to enhance feasibility and performance. \changes{Additionally, we evaluate D-MPC under the same setting. For a fair comparison, we train its planner and dynamics models following the original formulation~\citep{zhou2025diffusion}, using architectures comparable to ours. We find that adding more data does not improve D-MPC’s performance, as its dynamics model is applied only after planning for candidate filtering and does not influence action proposals. Consequently, a better dynamics model merely refines selection rather than generation. In contrast, MPDiffuser incorporates the dynamics model directly into the sampling process, allowing its improvements to immediately enhance the quality of generated trajectories.}

\begin{figure}[h]
    \centering
    \includegraphics[width=\linewidth]{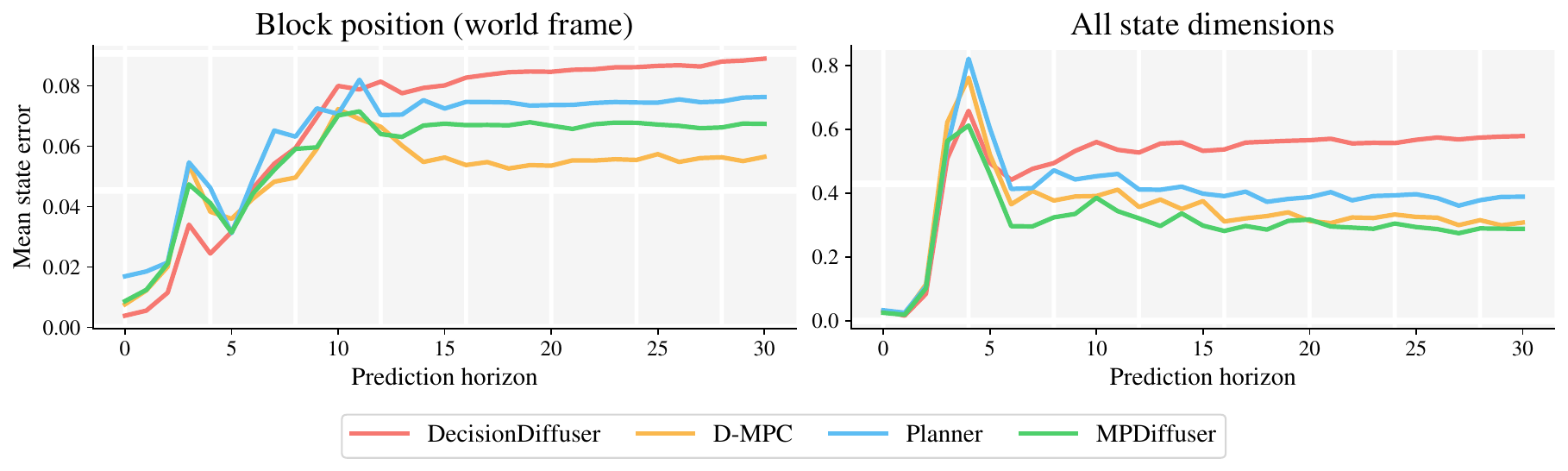}
    \caption{\textbf{Dynamics consistency of sampled trajectories.} Mean state error over the prediction horizon for block position and all state dimensions.}
    \label{fig:fetch-dynconsistency}
    \vspace{-10pt}
\end{figure}

\textbf{Assessing Dynamics Consistency of Sampled Trajectories:} 
We evaluate the dynamics consistency of trajectories generated by different diffusion models on the \texttt{FetchPickAndPlace} environment. Each model is trained on $1,000$ expert demonstrations. For evaluation, we sample $250$ random initial states, generate trajectories using each model, and compare the simulated rollouts (obtained by executing the sampled actions) with the diffused state trajectories. The mean errors are reported in Fig.~\ref{fig:fetch-dynconsistency}. MPDiffuser demonstrates stronger dynamics consistency than both Decision Diffuser and the planner-only baseline, while D-MPC achieves a comparable performance. However, despite the similar state deviation, MPDiffuser achieves a notably higher success rate (75\%) than D-MPC (60\%) as noted in Table~\ref{tab:fetch}, highlighting that our alternating sampling scheme balances task fidelity with dynamic feasibility more effectively. This difference arises because MPDiffuser integrates the dynamics model directly into the sampling process, actively correcting trajectories at each diffusion step, whereas D-MPC applies dynamics only as a post-hoc filtering.

\textbf{Adapting to Novel Dynamics:} To assess our method’s adaptability to changing system dynamics, we follow the experimental protocol from~\citet{zhou2025diffusion}. Accordingly, we train models on the D4RL \texttt{walker2d-medium} dataset and simulate a hardware defect by limiting the torque of one ankle joint to the range $[-0.5, 0.5]$. Table~\ref{tab:defect} summarizes the results, for Diffuser and D-MPC results are obtained from~\citet{zhou2025diffusion}. Originally, all methods achieve similar returns; yet, when deployed under the defect, both Diffuser and D-MPC suffer substantial performance drops, while MPDiffuser maintains significantly higher returns.

\begin{table}[h!]
  \begin{minipage}[t]{0.15\textwidth}
    \vspace{0pt}
    \centering
    \includegraphics[width=0.45\linewidth]{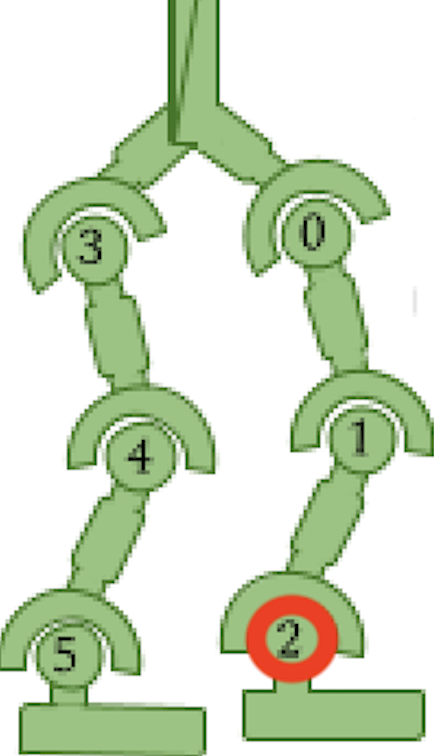}
    \captionsetup{type=figure}
    \captionof{figure}{Walker2D illustration, higlights defective joint}
  \end{minipage}
  \hfill
  \begin{minipage}[t]{0.3\textwidth}

    \vspace{0pt}
    \centering
      \small
\resizebox{\textwidth}{!}{
\begin{tabular}{lccc}
\toprule
 & Original & Pre-FT & Post-FT \\
\midrule
Diffuser & 79.6 & 25.9 & 6.8 \\
D-MPC & 76.2 & 22.7 & 30.7 \\
\changes{Planner} & \changes{75.9} & \changes{58.6} & \changes{56.0} \\
MPD & 77.6 & 58.6 & 66.4 \\
\changes{MPD+Rank} & \changes{77.6} & \changes{51.0} & \changes{63.4} \\
\bottomrule
\end{tabular}}
\caption{\textbf{MPDiffuser can adapt to novel dynamics.} Performance before and after fine-tuning (FT) under defect. }
    \label{tab:defect}
  \end{minipage}
  \vspace{-20pt}
\end{table}

\begin{table*}[h!]
\centering
\resizebox{\textwidth}{!}{\begin{tabular}{l|cc cc cc cc cc cc cc}
\toprule
 & \multicolumn{2}{c}{BC-All} 
 & \multicolumn{2}{c}{COptiDICE} 
 & \multicolumn{2}{c}{BC-Safe} 
 & \multicolumn{2}{c}{CPQ} 
 & \multicolumn{2}{c}{BCQ-Lag} 
 & \multicolumn{2}{c}{CDT} 
 & \multicolumn{2}{c}{MPDiffuser}\\
 & Return & Cost & Return & Cost 
 & Return & Cost & Return & Cost & Return & Cost 
 & Return & Cost & Return & Cost \\
\midrule

HopperVelocity 
& \textcolor{gray}{0.65} & \textcolor{gray}{6.39} 
& \textcolor{gray}{0.13} & \textcolor{gray}{1.51} 
& 0.36 & 0.67 
& \textcolor{gray}{0.14} & \textcolor{gray}{2.11}
& \textcolor{gray}{0.78} & \textcolor{gray}{5.02}
& 0.63 & 0.61  
& \textbf{0.81} & \textbf{0.37}\\

Walker2dVelocity 
& \textcolor{gray}{0.79} & \textcolor{gray}{3.88} 
& 0.12 & 0.74 
& \underline{0.79} & \underline{0.04} 
& 0.04 & 0.21
& \underline{0.79} & \underline{0.17}
& \underline{0.78} & \underline{0.06} 
& \textbf{0.80} & \textbf{0.27}\\

HalfCheetahVelocity 
& \textcolor{gray}{0.97} & \textcolor{gray}{13.1} 
& 0.65 & 0.0 
& 0.88 & 0.54 
& 0.29 & 0.74 
& \textcolor{gray}{1.05} & \textcolor{gray}{18.21}
& \textbf{1.0} & \textbf{0.01} 
& \underline{0.98} & \underline{0.77}\\ 

\midrule

PointGoal1 
& 0.65 & 0.95 
& \textcolor{gray}{0.49} & \textcolor{gray}{1.66} 
& 0.43 & 0.54 
& 0.57 & 0.35
& \underline{0.71} & \underline{0.98}
& \textcolor{gray}{0.69} & \textcolor{gray}{1.12}  
& \textbf{0.74} & \textbf{0.88} \\

PointCircle1 
& \textcolor{gray}{0.79} & \textcolor{gray}{3.98} 
& \textcolor{gray}{0.86} & \textcolor{gray}{5.51} 
& 0.41 & 0.16 
& 0.43 & 0.75
& \textcolor{gray}{0.54} & \textcolor{gray}{2.38}
& \textbf{0.59} & \textbf{0.69} 
& \underline{0.58} & \underline{0.94}\\

CarGoal1 
& 0.39 & 0.33 
& 0.35 & 0.54 
& 0.24 & 0.28 
& \textcolor{gray}{0.79} & \textcolor{gray}{1.42}
& 0.47 & 0.78
& \textcolor{gray}{0.66} & \textcolor{gray}{1.21}  
& \textbf{0.63} & \textbf{0.92} \\

CarCircle1 
& \textcolor{gray}{0.72} & \textcolor{gray}{4.39} 
& \textcolor{gray}{0.70} & \textcolor{gray}{5.72} 
& \textcolor{gray}{0.37} & \textcolor{gray}{1.38} 
& \textcolor{gray}{0.02} & \textcolor{gray}{2.29} 
& \textcolor{gray}{0.73} & \textcolor{gray}{5.25} 
& \textcolor{gray}{0.60} & \textcolor{gray}{1.73}  
& \textbf{0.50} & \textbf{0.85}\\
\bottomrule
\end{tabular}}
\caption{\textbf{Performance on DSRL benchmark tasks.} Normalized returns and costs on the DSRL benchmark, with gray entries indicating unsafe behavior. MPDiffuser achieves competitive returns while maintaining safety, demonstrating effective safety-performance balance.}
\label{tab:dsrl}
\vspace{-15pt}
\end{table*}

To adapt to the new dynamics, we collect 100 episodes of “play” data using our policy and fine-tune only the dynamics diffusion model. After fine-tuning, D-MPC partially recovers performance, whereas Diffuser further deteriorates. In contrast, MPDiffuser substantially improves and achieves the highest post-finetuning performance, confirming that isolating and updating the dynamics model allows efficient adaptation. The planner shows a slight decrease in performance after fine-tuning, suggesting that the absence of a dynamics component, causes it to forget previously learned behavior rather than adapt to new dynamics. Interestingly, MPDiffuser+Rank initially performs worse after the defect due to the ranker’s stronger bias toward high-return trajectories, which amplifies distribution shift under changed dynamics. Nevertheless, fine-tuning only the ranker suffices to recover its performance, demonstrating that both modules can be adapted independently and efficiently.

\subsection{Constrained Offline Decision Making}\label{sec:exp-const-offline}

\textbf{Results on Standardized Benchmarks (DSRL):} We evaluate our method on the DSRL benchmark~\citep{dsrl}, which includes safety-critical velocity and Safety Gym tasks. The objective is to maximize return while keeping cumulative cost below a specified budget. We compare against behavior cloning (BC-All),  behavior cloning trained only on safe trajectories (BC-Safe), cost-regularized approaches COptiDICE~\citep{lee2022coptidice}, CPQ~\cite{xu2022constraints}, BCQ-Lag~\cite{xu2022constraints}, as well as transformer-based CDT~\citep{liu2023constrained}. Our method, MPDiffuser, is tested  using 16 samples with learned cost and reward functions parameterized as MLPs. For each task, all methods are evaluated under cost budgets of $20$, $40$, and $80$ reporting average normalized return and cost over $60$ trials per budget. 

Table~\ref{tab:dsrl} shows that MPDiffuser consistently achieves high returns while adhering to the cost constraints. Notably, by varying the cost and return scale parameters during evaluation, the same trained model can flexibly generate behaviors across a wide spectrum of safety–reward tradeoffs, demonstrating the ability of our framework to adapt to diverse safety requirements without retraining.

\textbf{Importance of Dynamic Feasibility for Ranker:}  We  evaluate our approach on the classic \texttt{Pendulum} environment, modified with a hard velocity constraint requiring angular velocity to remain below $6.5\,\mathrm{m/s}$. We first train a standard soft actor-critic (SAC)~\citep{haarnoja2018soft} agent for $100$k steps, which frequently violates the velocity constraint, and a safe SAC agent that penalizes constraint violations heavily. To construct the dataset, we use the replay buffer of the unsafe SAC agent and $300$ trajectories collected from the safe SAC agent. We then compare MPDiffuser  with only planner based sampling and \changes{SafeDiffuser~\citep{xiao2023safediffuser} enforces safety by projecting sampled states within constraints during the diffusion process using a barrier-function.}

\begin{figure}[t]
  \begin{minipage}[ht]{0.13\textwidth}
    \vspace{0pt}
    \centering
    \includegraphics[width=0.7\linewidth]{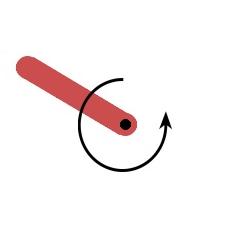} 
    \captionsetup{type=figure}
    \captionof{figure}{Pendulum environment.}
  \end{minipage}
  \hfill
  \begin{minipage}[ht]{0.33\textwidth}
    \vspace{0pt}
    \captionsetup{type=table}
    \resizebox{\textwidth}{!}{
    \begin{tabular}{@{}ccccccc@{}}
      \toprule
        Num. Samples & 1 & 4 & 8 & 16 & 32 & 64 \\
      \midrule
      Planner     & 62 & 89 & 91 & 88 & 74 & 66 \\
      \changes{SafeDiffuser}       & \changes{49} & \changes{62} & \changes{47} & \changes{42} & \changes{46} & \changes{45} \\
      MPDiffuser     & 69 & 84 & 93 & 93 & 92 & 91 \\
      \bottomrule
    \end{tabular}}
    \captionof{table}{\textbf{Ranking without dynamic feasibility violates safety.} Success rate comparison for varying number of samples.}
    \label{tab:pendulum}
  \end{minipage}
  \vspace{-22pt}
\end{figure}

Table~\ref{tab:pendulum} shows success rates as a function of sampled trajectories, where success means stabilizing the pendulum upright without violating the velocity constraint. \changes{SafeDiffuser achieves substantially lower success rates than other methods, underscoring the necessity of dynamic feasibility for projection-based approaches to maintain safety}. While planner initially improves with more samples, its performance later drops significantly because it often generates dynamically infeasible rollouts. With more samples, the chance of selecting such “hallucinated” trajectories that appear high-return but fail in practice increases. In contrast, MPDiffuser sustains high success rates as samples increase, highlighting robustness from our alternating sampling scheme.

\begin{figure*}[t]
  \centering
  \begin{minipage}[h]{0.5\textwidth}
    \centering
    \vspace{0pt}
    \includegraphics[width=\linewidth]{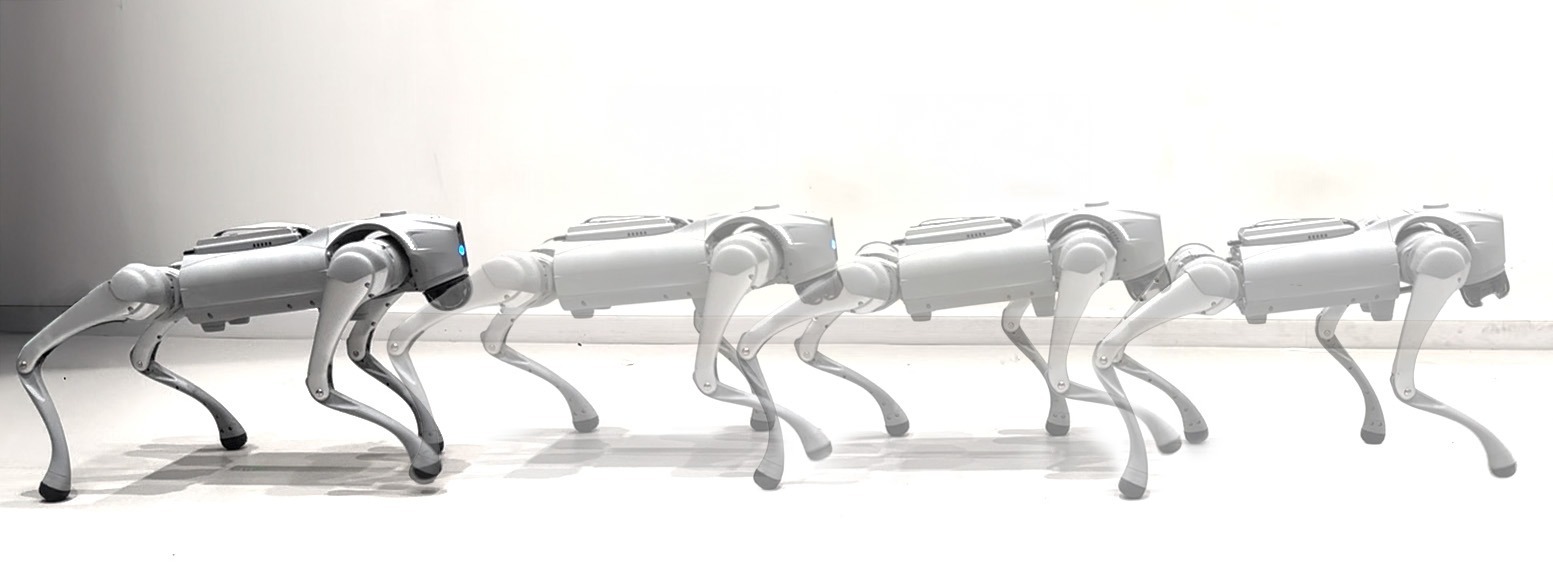}%
    \caption{Unitree Go2 quadruped robot walking.}
    \label{fig:go2-vel}
  \end{minipage}\hfill
  \begin{minipage}[h]{0.45\textwidth}
    \centering
    \includegraphics[width=\linewidth]{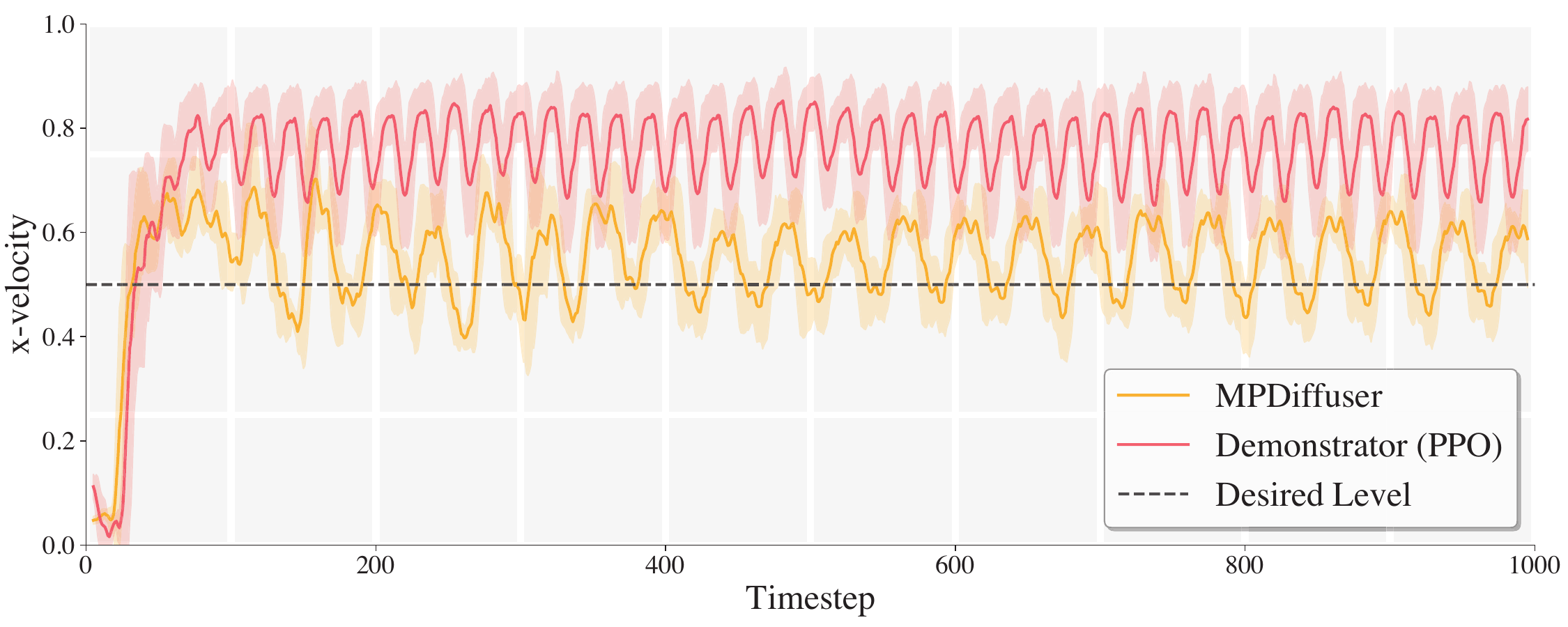}%
    \vspace{0pt}
    \caption{\textbf{Real-world demo.} Estimated velocity from the Unitree Go2 deployment.}
    \label{fig:go2-real}
  \end{minipage}
  \vspace{-10pt}
\end{figure*}

\vspace{-5pt}
\subsection{Extending MPDiffuser to Visual Domains}\label{sec:vision}
\vspace{-5pt}

To assess scalability to high-dimensional visual inputs, we conduct a preliminary proof-of-concept experiment on the Pendulum environment with image observations. We train a SAC agent for 100k transitions and use its replay buffer as the offline dataset. As observations we use centered grayscale images of the pendulum resized to $64\times64$, and stacked over four frames for temporal context.

\begin{table}[h!]
\centering
\resizebox{\columnwidth}{!}{
\begin{tabular}{l|cccc}
\toprule
 & Diffuser & Decision Diffuser & Planner & MPDiffuser \\
\midrule
Avg. Return & -196.2 & -242.9 & -181.5 & \textbf{-155.4} \\
\bottomrule
\end{tabular}}

\caption{\textbf{MPDiffuser scales effectively to visual inputs.} Average return over 250 evaluation trials.}
\label{tab:pendulum-img}
\vspace{-15pt}
\end{table}

We train a residual convolutional autoencoder to obtain a compact latent representation of these stacked frames with latent dimension 32. In addition to the standard reconstruction loss, we introduce a latent-space dynamics loss by training an auxiliary dynamics predictor that maps the current latent and action to the next latent. This encourages the learned representation to better reflect the system’s underlying dynamics. After training the autoencoder, we apply our MPDiffuser framework in this latent space and compare its performance against Decision Diffuser, Planner-only, and Diffuser baselines. As summarized in Table~\ref{tab:pendulum-img}, our approach achieves higher average returns, demonstrating that MPDiffuser can scale to visual domains and show superior performance even when operating on a learned latent space.

\vspace{-5pt}
\subsection{Robot Locomotion with Unitree Go2} \label{sec:exp-dog}
To assess real-world applicability, we evaluate our framework on quadruped locomotion using the Unitree Go2. Experiments are done in IsaacLab~\citep{mittal2023orbit} using configurations from the official Unitree repository. The state includes base angular velocity, projected gravity, and joint angles and velocities. Default domain randomization parameters is applied inducing stochasticity in dynamics. The reward promotes accurate velocity tracking via an exponentially decaying penalty on tracking error, while the cost activates when the gravity projection’s $z$-coordinate exceeds $-0.95$, encouraging parallel torso with a safety budget of $10$. A PPO~\citep{schulman2017proximal} policy is trained for $1000$ epochs to track constant velocity commands. As our dataset, we use $5000$ rollouts from this policy at four training snapshots (epochs $100$, $400$, $700$, and $1000$).

\begin{table}[h]
\centering
\resizebox{\columnwidth}{!}{
\begin{tabular}{l|cccc}
\toprule
 & Diffuser & Decision Diffuser & Planner & MPDiffuser \\
\midrule
Avg. Return & 74.7 & 84.9 & 94.7 & 94.8 \\
Cost   & 1.54 & 1.58 & 1.05 & 0.91 \\
\bottomrule
\end{tabular}}
\caption{\textbf{MPDiffuser matches performance while maintaining safety.} Performance of baseline methods and MPDiffuser on Unitree Go2 simulation. Returns and costs are normalized relative to the dataset average and cost budget, respectively.}
\label{tab:go2}
\vspace{-10pt}
\end{table}

In Table~\ref{tab:go2} we report average reward and cost computed over $1,250$ trials. MPDiffuser \ achieves the highest performance while remaining under the cost limit, 
highlighting the benefit of alternating planner–dynamics updates. In contrast, single-model baselines either generate 
unsafe trajectories or suffer from degraded returns due to their inability to balance task fidelity with dynamic feasibility. 
These results demonstrate that our compositional sampling strategy is crucial for reliable deployment in safety-critical locomotion tasks.

Finally, we validate our approach on the \textit{Unitree Go2} quadruped, running fully onboard with a Jetson Orin. Due to limited compute, we use single-sample inference and system-level optimizations (see Sec.~\ref{sec:apx-go2}). The robot tracks a constant velocity command of $[0.5, 0, 0]$, compared against a PPO policy trained for $1000$ epochs. Since direct velocity measurements are unavailable, a small MLP is trained to estimate velocity from joint states. As shown in Fig.~\ref{fig:go2-vel}, PPO overshoots the target, while MPDiffuser closely matches it. The PPO policy achieves $0.74\,\text{m/s}$, whereas MPDiffuser maintains $0.55\,\text{m/s}$, near the commanded $0.5\,\text{m/s}$. MPDiffuser attains a normalized return of $1.02$ versus $0.98$ for PPO, both with zero cost. Thus, this experiment confirms the practicality of MPDiffuser for real-world applications.

\vspace{-5pt}
\subsection{Analysis}
\label{sec:analysis}

\textbf{Effect of Alternating Planner--Dynamics Updates.} Due to our alternating sampling procedure we perform twice as many denoising steps as a planner-only model with the same number of steps. To ensure that the observed performance gains are not merely due to the increased number of updates, we compare: (i) Planner (100 steps), (ii) MPDiffuser (100 alternating steps), and (iii) Planner (200 steps).

Table~\ref{tab:alternate} reports average normalized returns across three D4RL medium-replay datasets. MPDiffuser (100 steps) achieves substantially higher returns than the planner-only variants, even when the planner is given twice as many denoising steps. The average computation time per action is 0.287 s for Planner (100), 0.575 s for Planner (200), and 0.583 s for MPDiffuser (100), evaluated using a single NVIDIA RTX 4090 GPU. Importantly, the runtime of MPDiffuser (100) is nearly identical to that of Planner (200), indicating that the observed performance gains arise from the compositional planner--dynamics sampling rather than from an increased number of diffusion steps.

\begin{table}[h]
\centering
\resizebox{\columnwidth}{!}{
\begin{tabular}{lccc}
\toprule
\textbf{Environment} & \textbf{Planner (100)} & \textbf{Planner (200)} & \textbf{MPDiffuser (100)} \\
\midrule
Hopper      & 92.1 & 89.9 & \textbf{98.2} \\
Walker2d    & 71.8 & 73.5 & \textbf{81.5} \\
HalfCheetah & 44.0 & 42.9 & \textbf{43.4} \\
\midrule
\textbf{Average} & 69.3 & 68.8 & \textbf{74.4} \\
\bottomrule
\end{tabular}}
\caption{\textbf{Alternating planner--dynamics sampling improves performance.}
Average normalized return across three D4RL medium-replay tasks. MPDiffuser consistently outperforms both planner-only variants, demonstrating the benefit of integrating a dynamics model within the sampling process.}
\label{tab:alternate}
\end{table}
\vspace{-10pt}

\textbf{Comparison of Alternation and Score Combination Methods.}
We study the effect of alternating versus combined score updates, motivated by Eq.~\eqref{eq:combined_score}. In the combined setting, we directly form a convex combination of the planner and dynamics scores over state dimensions at each diffusion step and perform a single denoising update, rather than alternating between the two models. We evaluate both approaches on the D4RL hopper environments. For the combined-score method, we perform a grid search over the weighting parameter and report the best result.

As shown in Table~\ref{tab:combined_vs_alt}, the combined-score update leads to consistently lower performance across all environments. We attribute this to gradient interference between the planner and dynamics components---since their objectives differ in curvature and scale, summing their scores produces unstable updates that can push trajectories away from feasible or high-reward regions. Alternating updates, in contrast, act as a form of operator splitting: each sub-step refines trajectories along a distinct objective, allowing the sampler to balance task fidelity and dynamics consistency more effectively.

\begin{table}[h]
\centering

\begin{tabular}{lcc}
\toprule
\textbf{Environment} & \textbf{Combined Score} & \textbf{Alternating} \\
\midrule
Medium-Expert & 106.9 & \textbf{110.4} \\
Medium        & 90.4  & \textbf{98.4} \\
Medium-Replay & 97.2  & \textbf{98.3} \\
\midrule
\textbf{Average} & 98.2 & \textbf{102.4} \\
\bottomrule
\end{tabular}
\caption{\textbf{Alternation outperforms combined updates.}
Normalized return on D4RL hopper environments. Alternating updates consistently outperform combined score updates, indicating that separate planner--dynamics denoising steps provide more stable and effective guidance.}
\label{tab:combined_vs_alt}
\end{table}
\vspace{-10pt}

\textbf{Limitations of MPDiffuser.}
While MPDiffuser is robust across all experiments where the planner and dynamics model are trained on the same dataset, we also investigate an intentionally mismatched setting to study potential failure modes. Specifically, we combine a planner trained on \texttt{medium-expert} data with a dynamics model trained on \texttt{medium-replay}. Although the replay dataset provides broader transition coverage, it lacks high-velocity expert demonstrations. As a result, the action proposals generated by the expert-trained planner fall partially outside the distribution seen by the replay-trained dynamics model, creating a distribution shift during the alternating updates.

This artificial mismatch leads to a notable drop in performance (Table~\ref{tab:limit}). In Hopper, the degradation is substantial: the ``Mixed'' MPDiffuser performs even worse than using a planner and dynamics model both trained on \texttt{medium-replay}. This suggests that, under sufficient mismatch, the dynamics module may over-correct trajectories toward its own training distribution, effectively harming performance. Importantly, this behavior does not appear in any of our main experiments, where both modules are trained on the same dataset---MPDiffuser remains stable and consistently improves over single-model baselines. Overall, this controlled failure case highlights a practical guideline rather than a fundamental limitation: MPDiffuser performs reliably when planner and dynamics modules are trained on compatible data distributions, which is the intended and natural usage of the framework.

\vspace{-20pt}
\begin{table}[h]
\centering
\begin{tabular}{lccc}
\toprule
\textbf{Environment} & \textbf{Med-Rep} & \textbf{Med-Exp} & \textbf{Mixed} \\
\midrule
Hopper      & 98.2 & 109.9 & 70.3 \\
Walker2d    & 81.5 & 110.7 & 81.7 \\
HalfCheetah & 43.4 & 96.9  & 49.0 \\
\bottomrule
\end{tabular}
\caption{\textbf{Effect of cross-dataset training.}
``Med-Rep'' and ``Med-Exp'' refer to MPDiffuser where both modules are trained on the same dataset; ``Mixed'' uses a planner trained on medium-expert and a dynamics model trained on medium-replay.}
\label{tab:limit}
\end{table}

\vspace{-28pt}
\section{Discussion}
\vspace{-3pt}

We introduced \textit{Model Predictive Diffuser} (MPDiffuser), a model-based diffusion framework that composes planner, dynamics, and ranker modules to synthesize task-aligned and dynamically feasible behaviors from offline data. By interleaving planner and dynamics updates, our sampling scheme improves both fidelity to demonstrations and consistency with system dynamics, leading to state-of-the-art performance across unconstrained (D4RL) and constrained (DSRL) benchmarks, as well as real-world robotic deployment. While our focus has been on the offline setting, future work could explore extending MPDiffuser to online decision-making by leveraging the dynamics module for exploration or adaptive control. \changes{Another promising direction is scaling our framework to complex, high-dimensional sensory domains (e.g., vision-based control) by performing diffusion in  latent spaces similar to~\citet{xie2025latent}, as preliminarily demonstrated in Sec.~\ref{sec:vision}. Finally, we aim to extend the framework across multiple environments and augment the dynamics model with cross-domain data, similar in spirit to world models~\citep{ha2018world}.}

\section*{Acknowledgements}
This work was supported in part by the NSF AI Institute under Grant No.~2112085, NSF ECCS-2401390, and ONR Grant N000142512173. The authors also acknowledge support from the Kempner Institute, Pickle Robotics, Amazon AGI Labs, Meta Reality Labs, and Amazon Robotics. We also thank Runhan Huang for assistance with our hardware experiment.

\section*{Impact Statement}
\vspace{-3pt}

This paper presents work whose goal is to advance the field of machine learning. There are many potential societal consequences of our work, none of which we feel must be specifically highlighted here.

\bibliography{example_paper}
\bibliographystyle{icml2026}

\newpage
\appendix
\onecolumn
\def\changes{\textcolor{black}}

\section*{\Large \textbf{Appendix}}

In this appendix we provide additional experimental, architectural, and theoretical details to complement the main text. 
In Section~\ref{sec:hypm} we outline hyperparameter settings, model architectures, and visualizations of all benchmark environments. 
In Section~\ref{sec:car} we introduce the custom Car U-Maze navigation task that is used to generate our illustration Fig.~\ref{fig:car}. 
In Section~\ref{sec:condition} we compare two schemes for incorporating the initial state---inpainting versus FiLM-based conditioning---through an ablation on D4RL Hopper. 
In Section~\ref{sec:linsys} we consider a linear system with a stochastic expert, providing a controlled setting where feasibility can be studied in detail. 
\changes{
In Section~\ref{sec:robust}, we evaluate the performance of MPDiffuser under modeling errors in the dynamics model.
}
In Section~\ref{sec:comp-budget} we evaluate the trade-off between computation budget and replanning frequency, highlighting the efficiency of warm-started diffusion. 
In Section~\ref{sec:dyn-cond}, we analyze the effect of conditioning in the dynamics model, showing that incorporating task information improves overall performance and consistency.
In Section~\ref{sec:causal}, we examine whether adopting a causal architecture provides any performance benefit for MPDiffuser.
In Section~\ref{sec:params}, we study the effect of the guidance scale $w$ and the number of ranking samples on final performance.
In Section~\ref{sec:theory} we give a theoretical justification of our alternating planner–dynamics sampling procedure by formalizing it as an approximation to an exponential-tilted distribution. 
Finally, in Section~\ref{sec:apx-go2} we describe the implementation of our real-world deployment on the Unitree Go2 quadruped, including system-level optimizations for real-time planning.

\section{Related Work}\label{sec:relatedwork}

\textbf{Diffusion Model Based Control:} Diffusion models have recently been applied to a wide range of decision-making and control problems. Recent work such as \citet{pearce2023imitating, carvalho2023motion}, and \citet{luo2025generative} explored imitation learning and motion planning, showing that diffusion priors can generate smooth and diverse trajectories. In reinforcement learning, several approaches employ diffusion at the action level, where a single action is generated conditioned on the current state. For example, \citet{lu2023contrastive} introduce Q-guided sampling and demonstrate strong reward performance. However, complex tasks with constraints and multiple objectives often require reasoning over longer horizons. To this end, trajectory-level diffusion has been adopted in offline RL settings~\citep{janner2022planning, ajay2023conditional, he2023diffusion}, as well as for policy learning in robotics~\citep{chi2023diffusion,huang2025diffuse}. These methods underscore the flexibility of diffusion-based formulations, as trajectory-level modeling captures long-term dependencies, composes behaviors observed in data, and accommodates constraints more effectively than single-step action generation.

\changes{
\textbf{Dynamics-Aware Diffusion for Feasible Planning} While diffusion models can effectively capture the distribution of state–action trajectories, generating trajectories that are \emph{dynamically feasible} remains a fundamental challenge. Existing trajectory diffusion methods~\cite{janner2022planning, ajay2023conditional} synthesize rollouts directly in data space without enforcing the underlying system dynamics. As shown in follow-up studies~\cite{zhou2025diffusion} and corroborated by our results (Figure~\ref{fig:car}), this often yields trajectories that deviate from true transition structures—demonstrating that producing perfectly dynamically consistent sequences with diffusion models is inherently difficult.
Several recent works attempt to alleviate this issue through inverse dynamics models (IDMs). For instance, \citet{ajay2023conditional} first diffuse state sequences and infer actions via a learned IDM, but such trajectories are often unrealizable under true dynamics. Similarly, \citet{luo2025generative} employ a related strategy for long-horizon planning and report frequent failure cases due to imperfect inverse dynamics. In contrast, our framework never relies on a single inverse-dynamics mapping: we explicitly separate planning from feasibility and correct the state evolution at every diffusion step using a dedicated dynamics diffusion model. This eliminates the brittle dependence on IDMs and keeps the trajectory close to the dynamics manifold throughout sampling.
Safety-oriented extensions, such as \citet{umenberger}, project states onto constraint manifolds during sampling but rely on the unrealistic assumption of a perfect inverse dynamics model for safety guarantees. MPDiffuser avoids such assumptions entirely: feasibility is enforced by a learned dynamics model operating at each diffusion timestep, yielding trajectories that satisfy constraints because the underlying state evolution is kept consistent with system transitions—not because an idealized inverse model is assumed.
In the visual domain, \citet{xie2025latent} apply an IDM over latent representations obtained from autoencoders; however, the resulting latent dynamics are often not well-posed, leading to severe degradation in control performance (Sec.~\ref{sec:vision}). Our approach sidesteps this issue by maintaining a dedicated diffusion dynamics model even in latent space, ensuring that feasibility corrections remain well-defined and that visual rollouts do not drift into spurious latent transitions.
}

\textbf{Model Predictive Control and Diffusion-based Approximations:} MPC is a leading optimization-based framework valued for its ability to optimize objectives under explicit constraints over finite horizons~\citep{rawlings2017model}. Yet, solving its optimization online becomes intractable for complex models, intricate rewards, or nonconvex constraints. This has motivated \textit{approximate MPC}, where offline solutions are used to train surrogates that approximate MPC behavior more efficiently~\citep{hertneck2018learning}.
Diffusion models have recently emerged as powerful generative surrogates.~\citet{huang2024toward} show that they can approximate MPC solutions with near-global optimality. However, diffusion models lack feasibility guarantees, creating a gap between generated trajectories and realized ones~\citep{zhao2024long}.
Several works aim to close this gap.~\citet{zhou2025diffusion} propose D-MPC with disjoint models for actions and dynamics, while our method integrates planning and dynamics correction within each diffusion step, simultaneously enforcing feasibility and improving trajectory fidelity. \changes{By integrating the dynamics model directly into each diffusion step, MPDiffuser incorporates dynamics feedback \emph{during} sampling—whereas in D-MPC the dynamics model influences generation only indirectly through candidate ranking—so in the single-sample regime D-MPC’s dynamics model is effectively inert, while ours remains fully operational as an active component of the sampling process.}~\citet{romer2024diffusion} adopt an alternating scheme, projecting trajectories onto feasible manifolds after each planner step via explicit optimization. Yet such projections are ill-posed within the diffusion process, since forward process breaks dynamic consistency. By contrast, our dynamics diffusion model learns the dynamics-induced manifold at every diffusion timestep, enabling feasibility enforcement in a distributionally consistent way during generation.

\textbf{Physics-Informed Neural Networks.} Another related direction is physics-informed and dynamics-constrained learning methods, such as Physics-Informed Neural Networks (PINNs) \cite{raissi2019physics}, neural ordinary differential equations \cite{chen2018neural}, Hamiltonian and Lagrangian neural networks \cite{greydanus2019hamiltonian,cranmer2020lagrangian}, and differentiable physics-based control frameworks. These approaches incorporate physical structure directly into learning, either by penalizing violations of governing equations, embedding conservation laws into the architecture, or leveraging differentiable simulators during optimization. More broadly, constrained generative and control methods often enforce feasibility through explicit physics priors or constraint penalties \cite{nath2024application}. In contrast, our framework does not assume analytic access to the system equations, differentiable simulators, or known conservation laws. Instead, MPDiffuser learns dynamics consistency implicitly from offline trajectories through a dedicated diffusion dynamics model that is integrated directly into the sampling process via alternating planner--dynamics updates. This makes our approach applicable in settings where accurate analytical models are unavailable or difficult to specify, while still improving trajectory feasibility during generation. Bridging compositional diffusion-based planning with physics-informed and structure-preserving learning methods therefore represents an interesting direction for future work, particularly for safety-critical domains where both data-driven flexibility and strong physical consistency are essential.

\section{Hyperparameters and Model Architecture}\label{sec:hypm}
\begin{figure}[htbp]
  \centering
  \begin{subfigure}{0.24\textwidth}
    \includegraphics[width=\linewidth]{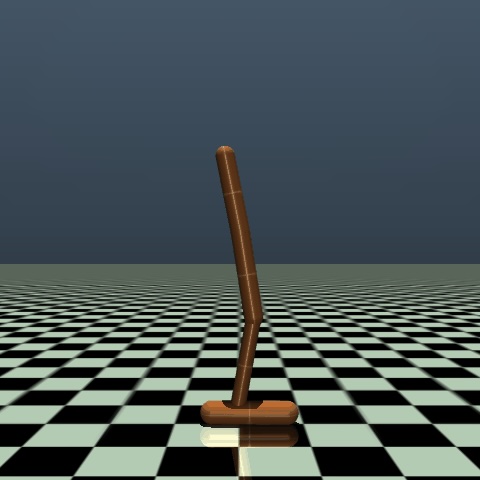}
    \caption{Hopper-v2}
  \end{subfigure}
  \begin{subfigure}{0.24\textwidth}
    \includegraphics[width=\linewidth]{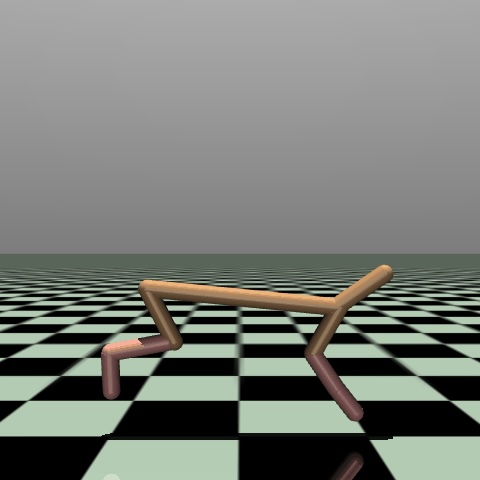}
    \caption{HalfCheetah-v2}
  \end{subfigure}
  \begin{subfigure}{0.24\textwidth}
    \includegraphics[width=\linewidth]{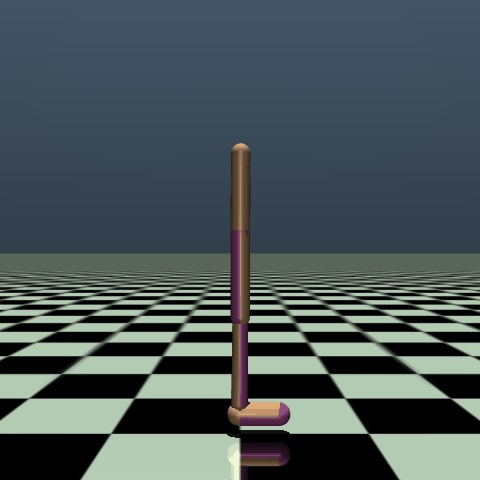}
    \caption{Walker2d-v2}
  \end{subfigure}
  \begin{subfigure}{0.24\textwidth}
    \includegraphics[width=\linewidth]{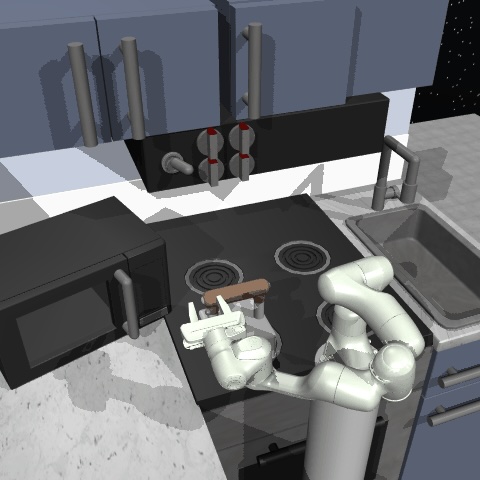}
    \caption{FrankaKitchen-v1}
  \end{subfigure}
  \caption{Datasets for Deep Data-Driven Reinforcement Learning (D4RL)~\citep{d4rl}.}
  
  \label{fig:first}
\end{figure}

\begin{figure}[htbp]
  \centering
  \begin{subfigure}{0.19\textwidth}
    \includegraphics[width=\linewidth]{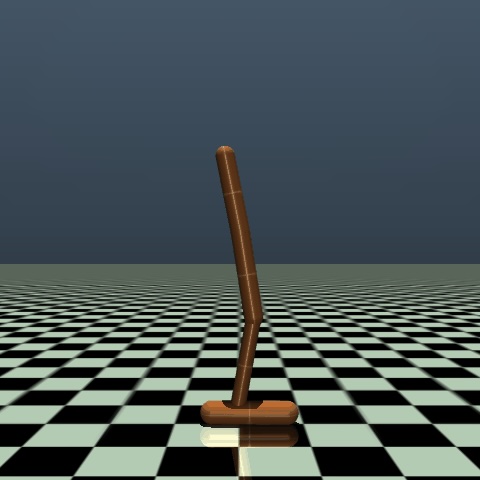}
    \caption{Hopper-v4}
  \end{subfigure}
  \begin{subfigure}{0.19\textwidth}
    \includegraphics[width=\linewidth]{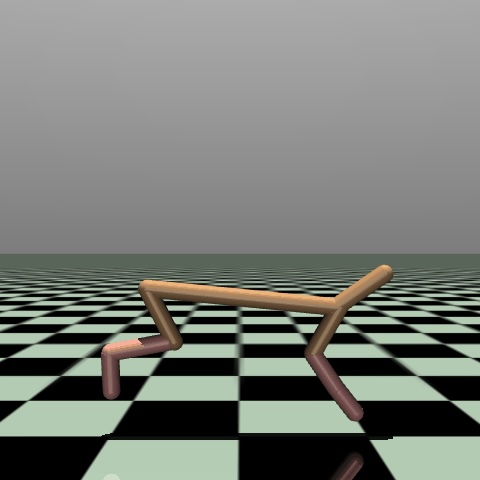}
    \caption{HalfCheetah-v4}
  \end{subfigure}
  \begin{subfigure}{0.19\textwidth}
    \includegraphics[width=\linewidth]{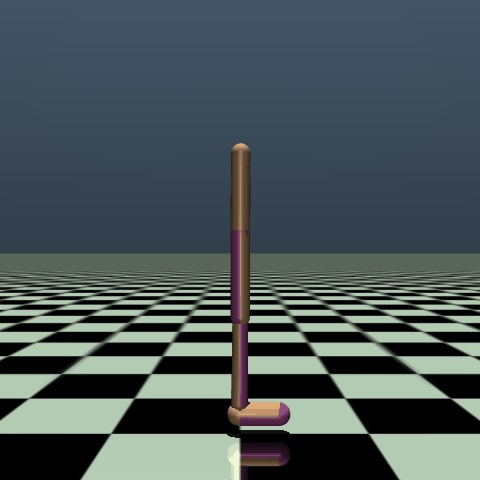}
    \caption{Walker2d-v4}
  \end{subfigure}
  \begin{subfigure}{0.19\textwidth}
    \includegraphics[width=\linewidth]{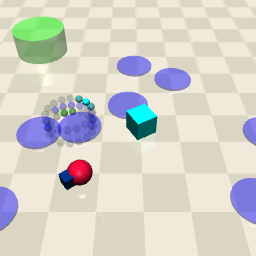}
    \caption{SafetyGymPoint}
  \end{subfigure}
  \begin{subfigure}{0.19\textwidth}
    \includegraphics[width=\linewidth]{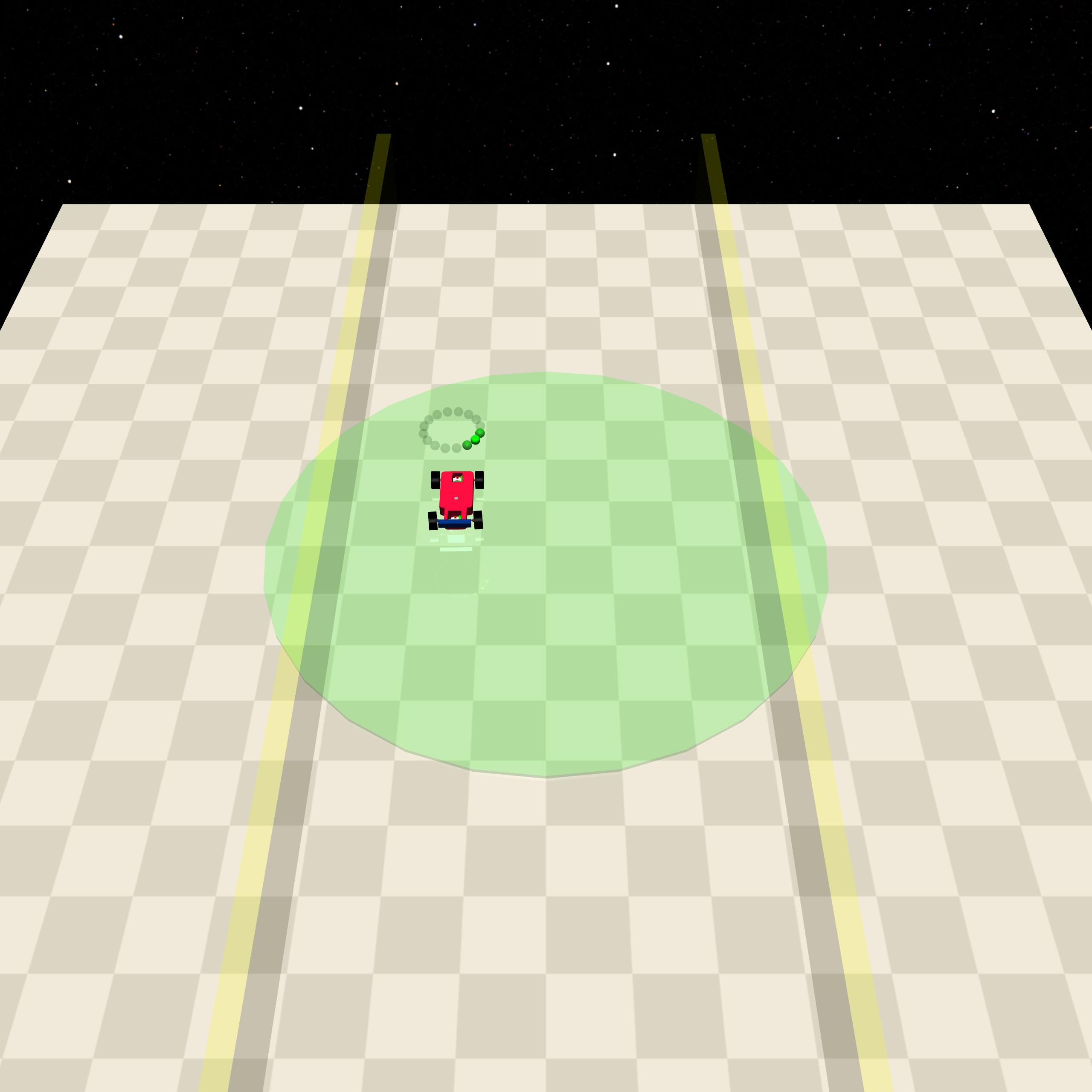}
    \caption{SafetyGymCar   }
  \end{subfigure}
  \caption{Datasets for Safe Reinforcement Learning (DSRL)~\citep{dsrl}}
  \label{fig:second}
\end{figure}

\begin{figure}[htbp]
  \centering
  \begin{subfigure}{0.19\textwidth}
    \includegraphics[width=\linewidth]{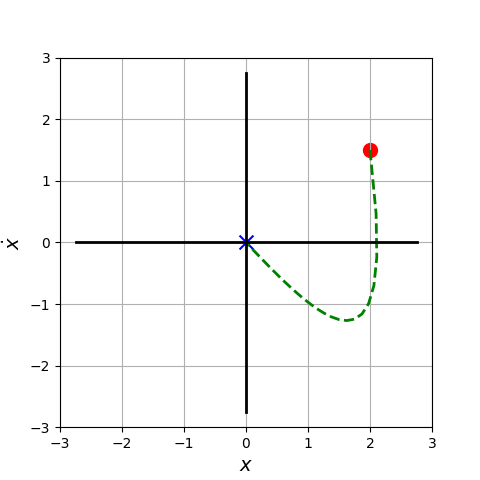}
    \caption{Double Integrator}
  \end{subfigure}
  \begin{subfigure}{0.19\textwidth}
    \includegraphics[width=\linewidth]{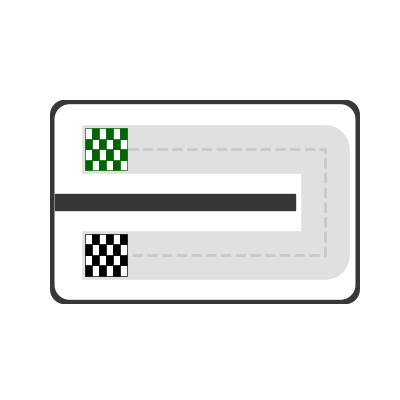}
    \caption{Car U-Maze}
  \end{subfigure}
  \begin{subfigure}{0.19\textwidth}
    \includegraphics[width=\linewidth]{images/env_imgs/custom/pendulum}
    \caption{Pendulum-v1}
  \end{subfigure}
  \begin{subfigure}{0.19\textwidth}
    \includegraphics[width=\linewidth]{images/env_imgs/custom/fetch}
    \caption{Fetch}
  \end{subfigure}
  \begin{subfigure}{0.19\textwidth}
    \includegraphics[width=\linewidth]{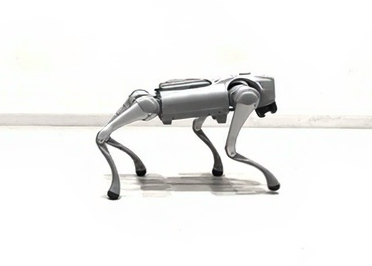}
    \caption{Unitree Go2}
  \end{subfigure}
  \caption{Custom datasets generated for this work.}
  \label{fig:third}
\end{figure}

In this section, we outline the key architectural and hyperparameter choices:

\begin{itemize}
    \item Both the planner noise model $\epsilon_\theta^{\mathrm{pl}}$ and the dynamics noise model $\epsilon_\theta^{\mathrm{dyn}}$ are implemented as temporal U-Nets as proposed by~\citet{janner2022planning}. Each network consists of six repeated residual blocks, where each block contains two temporal convolutions, followed by group normalization~\citet{wu2018group} and a Swish activation~\citet{swish}. Conditioning inputs $y(\tau)$ and the initial state $x_0$ are first processed with a two-layer MLP and then injected into the U-Net through FiLM layers~\citet{perez2018film}.
    
    \item We optimize $\epsilon_\theta$ and $f_\phi$ using Adam~\citep{kingma2014adam} with a learning rate of $2\times 10^{-4}$, a batch size of $64$, and $1 \times 10^6$ training steps. We track an exponential moving average of the weights with decay $0.005$, which is employed for evaluation.
    
    \item The conditioning vector is randomly dropped during training with probability $p=0.25$.
    
    \item We use $K=100$ diffusion steps for D4RL and DSRL benchmarks, $K=10$ for Unitree Go2 and $K=50$ steps for the remainder of custom datasets.
    
    \item The planning horizon is set to $H=64$ for D4RL Walker2d, DSRL Car EndPoint, and Pendulum environments, $H=16$ for the Unitree Go2 and $H=32$ for all other tasks.
    
    \item Guidance scale, return scale, temperature and cost scale are tuned separately for each task.

    \item The ranker module is a simple MLP with two hidden layers of 64 units and trained with mean-squared error with respect to rewards and costs.

\end{itemize}

\section{Car U-Maze}\label{sec:car}

We evaluate MPDiffuser on a custom navigation environment \texttt{CarU-Maze}, which requires navigating from a start position to a goal position using a 5-dimensional kinematic bicycle model. The training dataset is constructed by randomly sampling start-goal position pairs and generating corresponding U-shaped reference trajectories.  We collect $2000$ expert trajectories following the generated references using a nonlinear MPC controller, and additionally generate 1000 trajectories by sampling random actions to generate a diverse dataset.

For evaluation, we sample complete state-action trajectories from the trained diffusion models and execute the predicted actions in an open-loop manner within the environment. This open-loop execution enables direct comparison between the diffusion model's state predictions and the actual states that result from applying those actions under the true system dynamics. We assess performance using two complementary metrics: (1) the deviation between predicted and realized state trajectories, visualized qualitatively in Fig.~\ref{fig:car}, and (2) the Euclidean distance from the final achieved state to the target goal position.
This experimental setup demonstrates the capability of our compositional diffusion approach to generate trajectories that maintain dynamic consistency even under stringent kinematic constraints, highlighting its potential for complex control tasks requiring both geometric path planning and dynamic feasibility. We report success rates in Table~\ref{tab:success}, where a rollout is deemed successful if the Euclidean error between the final state and the target goal is less than $1.0$ units. As shown, MPDiffuser achieves success rates well above our baselines.

\begin{table}[h]
\centering
\begin{tabular}{lccc}
\toprule
 & Diffuser & Decision Diffuser & MPDiffuser  \\
\midrule
Success Rate (\%) & 68.8 & 42.2 & 95.3 \\
\bottomrule
\end{tabular}
\caption{\textbf{MPDiffuser achieves superior feasibility.} Success rates of different methods on the CarMaze task.}
\label{tab:success}
\end{table}

\section{Inpainting vs Conditioning}\label{sec:condition}
Most prior trajectory diffusion works (e.g., \citet{janner2022planning}) adopt a U-Net architecture that diffuses the entire sequence $(x_{0:T-1}, u_{0:T-1})$ and incorporates the initial state $x_0$ via an inpainting scheme. In contrast, we propose to inject $x_0$ directly through FiLM layers,~\citep{perez2018film}, while diffusing $(x_{1:T}, u_{0:T-1})$. This design ensures that the observed initial state is encoded consistently across the diffusion process without requiring partial trajectory masking. To evaluate the effect of this change, we conduct an ablation on D4RL Hopper tasks. As shown in Table~\ref{tab:d4rl-ablation}, conditioning through FiLM provides a consistent improvement over inpainting across all datasets, suggesting that our conditioning scheme is an effective way to incorporate initial state into trajectory diffusion models.  

\begin{table}[h]
\centering
\begin{tabular}{ll|cc}
\toprule
Dataset & Environment &  Inpainting & Conditioning\\
\midrule
\multirow{3}{*}{\makecell{Hopper}} 
& Med-Expert &  $108.1$ & $109.5$ \\
& Medium  &  $91.2$ & $97.6$   \\
& Med-Replay  & $87.4$ & $92.1$    \\
\midrule
& Average  & $95.6$ & $99.7$    \\
\bottomrule
\end{tabular}
\caption{\textbf{Ablation on initial state incorporation.} Comparison of inpainting versus FiLM-based conditioning on D4RL Hopper tasks. FiLM provides consistent improvements across datasets.}
\label{tab:d4rl-ablation}
\vspace{-10pt}
\end{table}

\section{Linear System with Stochastic Expert}\label{sec:linsys}

In this section, we consider a finite time optimal control problem of the form:
\begin{align}
    \min \;\; \sum_{t=0}^T \|x_t\|_Q^2 + \|u_t\|_R^2\quad 
    \textrm{s.t.}\;\; x_{t+1} = A x_t + B u_t,
\end{align}
where $A$, $B$ define the underlying linear time-invariant system and $\|x\|_Q^2 = x^\top Q x$ and $\|u\|_R^2 = u^\top R u$ define the quadratic cost function to be minimized. The system matrices $A$ and $B$ are derived from standard continuous time double integrator with sampling time $0.1\,\mathrm{s}$, and quadratic cost weights are set to be $Q = I$ and $R = 10^{-1} I$.

In the infinite-horizon case ($T \to \infty$), the optimal feedback controller is obtained by solving the discrete-time algebraic Riccati equation (DARE). Accordingly, the input and state trajectories under optimal control law can be computed as:
\begin{equation}
    u_t^\ast = K x^\ast_t,\quad  x_{t+1}^\ast =(A+BK)x^\ast_t, \quad K = \mathrm{DARE}(A, B, Q, R).
\end{equation}

Training data is generated from the optimal controller with additive Gaussian noise injected into the control input with probability $p$:
\begin{equation}
    u_t^{\mathrm{data}} = K x_t + d_t w_t, 
    \quad d_t \sim \mathrm{Bernoulli}(p), 
    \quad w_t \sim \mathcal{N}(0, 0.25^2 I),
\end{equation}
where the noise is i.i.d. across time. Both the training and evaluation phases use trajectories of length $200$, and models are trained on $1000$ trajectories.
In the training phase, the models are conditioned on the final state of each trajectory during the denoising process. In the evaluation phase, we generate sequences with the target final state fixed at the origin. 

\begin{figure}[t]
    \centering
    \includegraphics[width=0.8\linewidth]{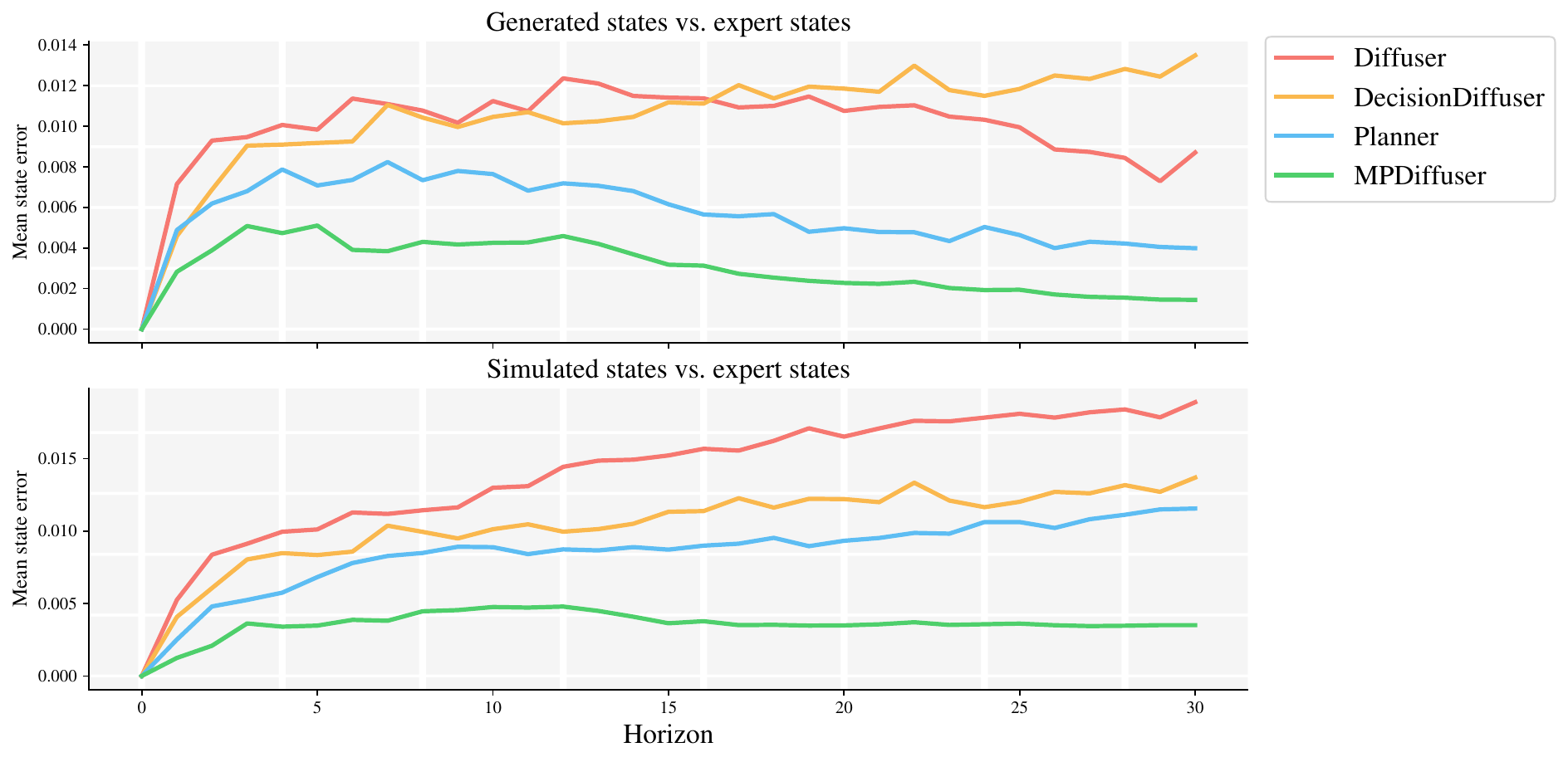}
    \caption{\textbf{MPDiffuser more closely aligns with expert behavior.} Average state error relative to the expert trajectory for $p=0.8$. The top panel compares generated (predicted) states from each method to the states obtained by expert. The bottom panel compares states obtained by simulating the system with the generated actions. The proposed method achieves the lowest error in both cases, highlighting the benefit of dynamics-consistent correction during generation.}
    \label{fig:linsys}
\end{figure}

\begin{table}[t]
    \centering
    \begin{tabular}{c|ccccc}
        \toprule
        \textbf{Noise Level} & \textbf{Diffuser} & \textbf{DecisionDiffuser} & \textbf{Planner} & \textbf{MPDiffuser} \\
        \midrule
        $p=0.1$ & 2.38 & 2.34 & 1.36 & 1.27 \\
        $p=0.2$ & 3.50 & 3.12 & 2.63 & 1.54 \\
        $p=0.3$ & 4.40 & 5.48 & 4.40 & 3.38 \\
        $p=0.4$ & 5.27 & 5.59 & 5.00 & 3.99 \\
        \bottomrule
    \end{tabular}
    \caption{\textbf{MPDiffuser is more robust to stochasticity in the data.} Performance comparison on the linear system example for different noise injection probabilities $p$. The cost values are normalized by average cost incurred under infinite-horizon optimal controller ($u=Kx$).}
    \label{tab:linear}
\vspace{-20pt}
\end{table}

In Table~\ref{tab:linear}, we report average cumulative costs computed over $250$ evaluation trials. As shown, the proposed method consistently achieves the lowest cost across all noise levels. The performance gap to the baselines widens as $p$ increases, i.e., when the dataset contains higher diversity and trajectories are further away to the optimal policy. In the high-noise regime, the demonstrations are highly suboptimal, making it difficult for standard diffusion models to synthesize trajectories close to the optimal evolution. However, even in this setting, the demonstrations remain dynamically consistent, providing the dynamics model with rich structure to exploit. As a result, the proposed approach’s dynamics-consistent correction step preserves feasibility during generation and yields improved performance despite the suboptimality of the data. When compared to the Planner model without dynamics correction, the proposed method yields significant improvements at lower noise levels, highlighting the importance of incorporating dynamics consistency during sampling.

To further analyze performance, we examine the deviation between generated state sequences and those produced by the optimal policy. For this experiment, we sample random initial states and generate state–action sequences for each method. Figure~\ref{fig:linsys} reports the average state error relative to the expert trajectory for dataset generated with policy noise level $p=0.8$ both for the generated (diffused) states and for states obtained by simulating the system with the generated actions. The results show that Diffuser and Decision Diffuser fail to produce accurate state sequences, while the Planner alone achieves moderate accuracy. MPDiffuser consistently achieves the lowest error in both settings. Moreover, when simulated using the generated actions, Diffuser and planner incur substantially higher errors than our method, indicating that our dynamics-consistent correction step improves not only quality of sampled trajectories but also open-loop performance under the generated actions.
\vspace{-1em}
\changes{
\section{Robustness to Dynamics Model Errors}\label{sec:robust}
The accuracy of the dynamics model is critical for the performance of MPDiffuser. To evaluate the robustness of our framework to modeling errors, we conduct an ablation where the dynamics model is trained on corrupted data with varying levels of state measurement noise. Specifically, we use the dataset corresponding to noise probability $p=0.8$ from the linear system setup (Sec.~\ref{sec:linsys}) and keep the planner fixed. The dynamics model is trained on the same dataset with additive Gaussian noise applied to the state measurements at different standard deviations. We then evaluate the resulting MPDiffuser models under each setting.}

\changes{As shown in Table~\ref{tab:dyn-noise}, increasing the level of corruption in the dynamics model leads to only a degradation in performance, demonstrating that MPDiffuser is robust to moderate modeling errors. Notably, even when the dynamics model is trained with substantial state noise, MPDiffuser continues to outperform the planner-only and other diffusion-based baselines. However, as the noise level increases the dynamics model quality drops further and eventually overall performance drops significantly. This study highlights the stabilizing role of alternating planner–dynamics updates, which preserve high task-fidelity rollouts even under imperfect dynamics estimation.}

\begin{table}[h]
\centering
\captionsetup{labelfont={color=black,bf}, textfont={color=black}}
\begin{tabular}{l|cccccccc}
\toprule
\textbf{Noise Std. ($\sigma$)} & 0.000 & 0.001 & 0.002 & 0.003 & 0.004 & 0.005 & 0.010 & 0.020 \\
\midrule
\textbf{Cost $\downarrow$} & 1.54 & 1.97 & 2.08 & 2.19 & 2.33 & 2.56 & 3.67 & 5.25 \\
\bottomrule
\end{tabular}
\caption{\textbf{Robustness to dynamics model errors.} Normalized cost (with respect to LQR controller) on the linear system dataset ($p=0.8$) when training the dynamics model with varying levels of measurement noise.}
\label{tab:dyn-noise}
\vspace{-20pt}
\end{table}

\section{Computation Budget, Replanning Experiment}\label{sec:comp-budget}

\begin{figure}[h]
    \centering
    \includegraphics[width=0.75\linewidth]{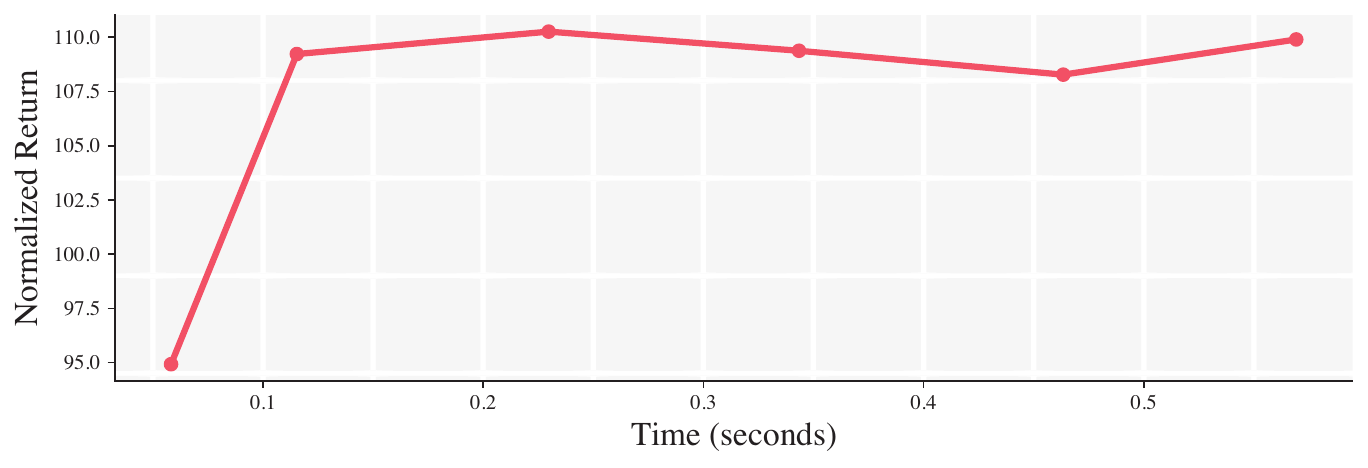}
    \caption{\textbf{Performance vs.~planning time:} Trade-off between performance (normalized average return), and planning cost, measured in wall-clock time after warm-starting the reverse diffusion process. The results are obtained using a single NVIDIA RTX 4090 GPU}
    \label{fig:comp-budget}
\vspace{-30pt}
\end{figure}

We analyze the runtime characteristics of our compositional diffusion procedure in the D4RL \texttt{hopper-medium-expert-v2} environment. After training both the planner and dynamics diffusion models, we generate trajectories according to Algorithm~\ref{alg:alt-guided}. Naively, each environment step requires running a full reverse diffusion chain, which can be computationally expensive. To accelerate planning, we adopt a warm-start strategy: the generated trajectory from the previous step is partially diffused forward for a fixed number of steps, after which the same number of reverse diffusion steps are applied to obtain a new trajectory as proposed in~\citep{janner2022planning}. Owing to the improved dynamic feasibility of our generated sequences, the warm-started trajectory remains close to the optimal continuation, since the observed environment state is typically very close to the predicted next state. As shown in Figure~\ref{fig:comp-budget}, the number of denoising steps can be reduced substantially with little loss in performance: using only $10$ steps yields an average return of $94.9$ with $58$\,ms per action, while beyond $20$ steps performance is comparable to running the full diffusion chain.

\changes{
\section{Impact of Conditioning on Dynamics Learning}\label{sec:dyn-cond}
In this section, we investigate the effect of conditioning in the dynamics diffusion model. Although forward dynamics are typically unconditional, conditioning the dynamics model on task or goal information can improve optimization stability and facilitate generation of high-reward trajectories. Within our alternating sampling framework, the planner drives trajectories toward task-specific objectives; an entirely unconditional dynamics model may weaken this coupling and hinder task alignment.}

\changes{
To evaluate this, we train both conditional and unconditional variants of the dynamics model on D4RL medium-replay environments while keeping the planner identical. As shown in Table~\ref{tab:cond-vs-uncond}, the conditional dynamics model consistently achieves higher normalized returns across tasks. These results suggest that conditioning provides beneficial structure for guiding feasible, task-relevant rollouts without sacrificing generality.
}

\begin{table}[h]
\centering
\color{black}
\captionsetup{labelfont={color=black,bf}, textfont={color=black}}
\begin{tabular}{l|cc}
\toprule
\textbf{Environment} & \textbf{Unconditional} & \textbf{Conditional} \\
\midrule
Hopper & 91.3 & \textbf{98.2} \\
Walker2d & 78.6 & \textbf{81.5} \\
HalfCheetah & 43.3 & \textbf{43.4} \\
\midrule
Average  & $71.1$ & $74.4$    \\
\bottomrule
\end{tabular}
\caption{\textbf{Dynamics model benefits from conditioning.} Normalized return on D4RL medium-replay tasks.}
\label{tab:cond-vs-uncond}
\end{table}
\vspace{-20pt}
\changes{
\section{Should Trajectory Denoisers Be Causal?}\label{sec:causal}
A natural question is whether the denoiser should mirror the forward-time causality of the underlying dynamics or whether such a restriction limits its modeling capacity.  
Motivated by this, and following observations in~\citet{chen2024diffusion}, we examine the effect of enforcing temporal causality in our denoising networks.  
We re-implement both the planner and dynamics models using causal U-Nets in the WaveNet style~\cite{rethage2018wavenet} and evaluate them on D4RL medium-replay tasks.
}

\changes{
As reported in Table~\ref{tab:causal-vs-acausal}, causal architectures lead to a slight drop in performance.  
Although system dynamics are inherently causal, the optimal denoiser in a diffusion model need not be: score estimation at each timestep is a smoothing operation that benefits from future context~\cite{wiener1964extrapolation}, and similar observations have been made in diffusion models for audio and speech~\cite{kong2020diffwave}.  
Our results align with this: view—strict causality restricts receptive fields and degrades the quality of the learned score, whereas acausal models exploit full-context information during denoising.
}

\begin{table}[h]
\centering
\color{black}
\captionsetup{labelfont={color=black,bf}, textfont={color=black}}
\begin{tabular}{l|cc}
\toprule
\textbf{Environment} & \textbf{Causal} & \textbf{Acausal} \\
\midrule
Hopper & 93.1 & \textbf{98.2} \\
Walker2d & 70.5 & \textbf{81.5} \\
HalfCheetah & 43.5 & \textbf{43.4} \\
\midrule
Average  & $69.0$ & $74.4$    \\
\bottomrule
\end{tabular}
\caption{\textbf{Acausal denoisers perform better.} Normalized return on D4RL medium-replay tasks.}
\label{tab:causal-vs-acausal}
\end{table}
\vspace{-2.5em}

\section{Parameter Sensitivity}\label{sec:params}
We evaluate the sensitivity of MPDiffuser to two key sampling hyperparameters on \texttt{FetchPickAndPlace}: the classifier-free guidance scale $w$, which controls the strength of task conditioning during denoising, and the number of sampled trajectories used by the ranker. As shown in Tables~\ref{tab:params1} and~\ref{tab:params2}, performance remains stable over a broad range of guidance strengths, with a mild peak around $w \in [1.5, 2.5]$. Increasing the number of samples for ranking yields improvements up to roughly 8 samples, after which returns saturate.

\begin{table}[h]
\centering
\color{black}
\captionsetup{labelfont={color=black,bf}, textfont={color=black}}
\begin{tabular}{l|ccccc}
\toprule
\textbf{CFG strength ($w$)} & 1.5 & 1.75 & 2.0 & 2.25 & 2.5 \\
\midrule
\textbf{Normalized Score} & 72 & 80.3 & 81.5 & 76.5 & 72.0 \\
\bottomrule
\end{tabular}
\caption{\textbf{Effect of classifier-free guidance scale.}  
Success rate on \texttt{FetchPickAndPlace}. MPDiffuser is robust to the choice of guidance strength.}
\label{tab:params1}
\end{table}

\begin{table}[h]
\centering
\color{black}
\captionsetup{labelfont={color=black,bf}, textfont={color=black}}
\begin{tabular}{l|cccccc}
\toprule
\textbf{Num. samples} & 1 & 2 & 4 & 8 & 16 & 32 \\
\midrule
\textbf{Normalized Score} & 60 & 62 & 70 & 73 & 75 & 72 \\
\bottomrule
\end{tabular}
\caption{\textbf{Effect of number of samples for ranking.}  
Success rate on \texttt{FetchPickAndPlace}. Performance stabilizes after $8$ samples, with a slight peak at $16$.}
\label{tab:params2}
\end{table}

\vspace{-0.5em}
\section{Theoretical Justification}
\label{sec:theory}
In this section, we provide theoretical justification for our algorithm by formulating trajectory generation as a constrained optimization problem that balances planner fidelity with dynamics feasibility. Our key insight is that the optimal sampling distribution can be characterized as an exponential tilting of the planner distribution, weighted by dynamics consistency. We show that while direct sampling from this distribution is intractable, our alternating update scheme offers a principled approximation inspired by operator splitting from numerical integration.

For notational simplicity, we omit explicit conditioning on trajectory conditioning vector $y(\tau)$, 
and write distributions as $p(\cdot \mid x_0)$ rather than $p(\cdot \mid x_0, y(\tau))$. 
All derivations can be simply extended to the conditioned case without any major modifications.

\paragraph{Defining dynamic feasibility.}
To measure whether a candidate trajectory $\tau = (x_{0:T},u_{0:T-1})$ is consistent with the system
dynamics, we define a trajectory likelihood under a dynamics-induced distribution. This distribution
factors into the conditional likelihood of the state sequence given the actions and a prior over the
actions themselves:
\begin{align}
p_{\mathrm{dyn}}(\tau \mid x_0)
&=\; \prod_{t=0}^{T-1} p_{\mathrm{dyn}}(x_{t+1}\mid x_t,u_t)\;\;p_{\mathrm{dyn}}(u_t).
\end{align}
In our setting, the conditional state transitions follow the system kernel, so we can write
\begin{align}
p_{\mathrm{dyn}}(\tau \mid x_0)
&=\; \prod_{t=0}^{T-1} P(x_{t+1}\mid x_t,u_t)\;\;p_{\mathrm{dyn}}(u_t).
\end{align}
Finally, to simplify the formulation, we assume that the dynamics distribution places equal
probability on all possible action realizations (i.e.\ $p_{\mathrm{dyn}}(u_t)$ is uniform). Under this
assumption the action prior contributes only a constant factor, which we drop, leading to
\begin{align}
p_{\mathrm{dyn}}(\tau \mid x_0)
\;\propto\;\prod_{t=0}^{T-1} P(x_{t+1}\mid x_t,u_t).
\end{align}
Thus $p_{\mathrm{dyn}}$ evaluates a trajectory based solely on how well its state sequence aligns
with the system dynamics, regardless of which particular actions are chosen.
The defined distribution assigns higher probability to the trajectories that are more probable under the transition kernel, while implausible trajectories are assigned lower probability. 
In the deterministic setting, the transition kernel reduces to a Dirac measure 
$P(x_{t+1} \mid x_t, u_t) = \delta(x_{t+1} - f(x_t, u_t))$. 
While this enforces strict feasibility by assigning nonzero probability only to the exact successor 
state, such a formulation is brittle in practice and precludes comparing trajectories that deviate 
even slightly from the dynamics. 
To address this, one often considers a relaxed kernel such as a Gaussian centered at the deterministic next state $f(x_t, u_t)$ with a desired level of variance. 
This yields a dense measure of trajectory quality: transitions closer to the dynamics model incur smaller penalties, while larger deviations are increasingly penalized. 
Under this relaxation, the dynamics log-probability reduces to a quadratic form similar to the 
squared-residual surrogate introduced earlier.

\paragraph{Defining planner distribution.}
Let $p_{\mathrm{pl}}(\tau  \mid x_0)$ denote the \emph{induced} trajectory distribution obtained by running a fixed (e.g., DDIM) sampling procedure from the learned score/denoiser, conditioned on the initial state $x_0$. Intuitively, $p_{\mathrm{pl}}$ concentrates on trajectories that resemble the dataset and thus capture task structure, and preferences present in demonstrations.

\paragraph{Projection toward dynamics feasibility.}
While $p_{\mathrm{pl}}$ yields high‑quality trajectories, its samples need not be fully consistent
with the system dynamics. To explicitly encourage feasibility, we utilize the dynamics probability function
$p_{\mathrm{dyn}}$ (defined above via the transition kernel) and seek a
\emph{nearby} distribution $q(\cdot \mid x_0)$ whose trajectories have higher dynamics probability. We formalize ``nearby'' by constraining the Kullback–Leibler divergence to lie within a
small radius $\varepsilon > 0$:
\begin{equation}
\label{eq:i_proj_constraint}
\begin{aligned}
\min_{q} \quad & 
\mathbb{E}_{q}\!\left[ -\log p_{\mathrm{dyn}}(\cdot \mid x_0) \right] \\[4pt]
\text{s.t.} \quad & 
\mathrm{KL}\!\left(q(\cdot \mid x_0)\,\|\,p_{\mathrm{pl}}(\cdot \mid x_0)\right) \;\le\; \varepsilon.
\end{aligned}
\end{equation}
The constraint preserves fidelity to the planner---retaining its task‑relevant structure and sample
quality---while the objective steers probability mass toward trajectories that are more probable
under the dynamics (i.e., higher $p_{\mathrm{dyn}}$).
In this sense, \eqref{eq:i_proj_constraint} is a projection of $p_{\mathrm{pl}}$ onto the set of
dynamics‑consistent distributions within a KL ball, yielding a principled balance between
\emph{planner fidelity} and \emph{dynamics feasibility}.

The constrained projection \eqref{eq:i_proj_constraint} can be handled via a Lagrangian relaxation,
leading to the unconstrained objective
\begin{equation}
\label{eq:i_proj_objective}
\min_{q}\;\; 
\mathcal{F}_\lambda(q)
:= \underbrace{\mathbb{E}_{q}\!\left[-\log p_{\mathrm{dyn}}(\cdot| x_0)\right]}_{\text{dynamics consistency}}
\;+\; 
\underbrace{\tfrac{1}{\lambda}\, \mathrm{KL}\!\big(q(\cdot \mid x_0)\,\|\,p_{\mathrm{pl}}(\cdot \mid x_0)\big)}_{\text{planner fidelity}},
\qquad \lambda > 0.
\end{equation}
Intuitively, $\lambda$ trades off fidelity to the planner against dynamics consistency: small $\lambda$ favors
$p_{\mathrm{pl}}$, while large $\lambda$ emphasizes high dynamics consistency.

\paragraph{Solution via Exponential Tilting.}
By Gibbs’ variational principle~\citet{cover1999elements}, the unique minimizer of \eqref{eq:i_proj_objective} is given by an exponential tilting of the planner distribution:
\begin{equation}
q^\ast(\tau \mid x_0)
\;\propto\;
p_{\mathrm{pl}}(\tau  \mid x_0)\;
\exp\!\big(\lambda \log p_{\mathrm{dyn}}(\tau \mid x_0)\big).
\end{equation}
Equivalently, we can write:
\begin{equation}
q^\ast(\tau \mid x_0)
\;\propto\;
p_{\mathrm{pl}}(\tau \mid x_0)\,
p_{\mathrm{dyn}}(\tau \mid x_0)^{\lambda}.
\end{equation}
Thus the optimal target distribution $q^\ast$ is a combination of the planner distribution and the dynamics distribution, with the exponent $\lambda$ controlling their relative influence.

\paragraph{Sampling from $q^\ast$.}
Directly characterizing $q^\ast$ is difficult in practice: we do not have an explicit form for the
planner distribution $p_{\mathrm{pl}}$ nor for the dynamics distribution $p_{\mathrm{dyn}}$, and
thus cannot evaluate or draw samples from their product-of-experts combination. An alternative is to
appeal to the diffusion framework, where one can sample from a target distribution by following a
discrete approximation of its probability--flow dynamics. At diffusion step $k$, DDIM update~\citet{song2020denoising} takes the form:
\begin{equation}
\tau^{k-1}
\;=\;\tau^k + f(\tau^k,k)\,\Delta_k \;-\; g(k)^2\,s_{q^\ast}(\tau^k,k)\,\Delta_k,
\end{equation}
where $\Delta_k$ is the effective step length defined by the noise schedule $\beta_k$ and
$s_{q^\ast}(\tau^k,k)=\nabla_\tau \log q^{\ast,k}(\tau^k)$ is the score of the corrupted marginal of
$q^\ast$ at noise level $k$. However, $q^\ast$ is only an abstract construction obtained by
combining the planner and dynamics distributions; we do not have direct samples from $q^\ast$. As a
result, we cannot directly train a diffusion model to estimate its score $s_{q^\ast}$.

\paragraph{Approximating the score of $q^\ast$.}
The exact score of the target distribution at diffusion step $k$ is:
\begin{equation}
s_{q^\ast}(\tau^k,k) = \nabla_{\tau^k} \log q^{\ast}_{k}(\tau^k) = \mathbb{E}_{\tau^0 \sim p(\tau^0 \mid \tau^k)}\!\left[\nabla_{\tau^k} \log q_{k\mid 0}^\ast(\tau^k \mid \tau^0)\right],
\end{equation}

where the expectation is over the posterior distribution of clean trajectories given the noisy observation. This expectation is intractable as it requires marginalizing over all possible clean trajectories consistent with $\tau^k$. Following common practice in score-based diffusion models, we approximate this with a sum of individual scores:
\begin{equation}
s_{q^\ast}(\tau^k,k) \approx s_{p^{\mathrm{pl}}}(\tau^k,k) + \lambda s_{p^{\mathrm{dyn}}}(\tau^k,k).
\end{equation}
This approximation is exact when the noise level approaches zero and becomes increasingly accurate for small noise levels typical in the later stages of sampling.

\changes{
\paragraph{Motivation for alternating updates.}
A natural way to approximate $s_{q^\ast}$ is to directly combine the planner and dynamics scores and perform a single joint update at each diffusion step. However, in practice this can lead to instability, as the planner and dynamics gradients often differ in scale, curvature, and local geometry—causing gradient interference that may push samples off-manifold. We empirically validate this observation in Sec.~\ref{sec:analysis}, where directly combining the scores results in consistently lower performance compared to our alternating update scheme. To mitigate this, our algorithm instead applies alternating planner and dynamics updates, each acting on a subset of variables while the other is held fixed. This separation yields more stable and interpretable behavior, allowing the dynamics model to enforce feasibility locally before the planner steers the trajectory toward higher reward regions.}

\changes{
This alternating procedure can be interpreted through the lens of \emph{fractional-step} or \emph{operator-splitting}~\citep{hairer2006geometric,trotter1959product,strang1968construction}. When a system evolves under two interacting vector fields---here represented by the planner and dynamics scores---alternating short integration steps under each component provides a first-order Lie–Trotter approximation to the joint flow. For sufficiently small step size, the global discretization error of the Lie–Trotter splitting decays linearly with step size. Hence, as the step size decreases, the alternating process converges to the true combined flow $s_{q^\ast}$.}

\changes{
Moreover, if the dynamics model provides a more accurate local estimate of the true dynamics score $s_{p^{\mathrm{dyn}}}$ than the planner, then alternating updates effectively correct the planner’s bias at each diffusion step. In our framework, the dynamics diffusion model generally provides a more accurate local approximation of the true dynamics score $s_{p^{\mathrm{dyn}}}$ than the planner. Although both modules share a twin network architecture and are trained on the same dataset, the dynamics model is specialized solely for state prediction, whereas the planner must jointly model both states and actions under task conditioning. This specialization allows the dynamics model to devote its capacity entirely to capturing transition consistency and the underlying physical structure of the environment. Empirically, we observe across all experiments that the dynamics model yields lower state-prediction error than the planner, which directly translates into improved feasibility when incorporated into the alternating sampling loop. Moreover, unlike the planner, the dynamics model can be trained effectively even on diverse or low-quality datasets, since it does not rely on optimal actions but only on accurate state transitions. This property is confirmed in our Sec.~\ref{sec:exp-offline}, where using additional random or suboptimal trajectories improves performance by enhancing the learned dynamics, further supporting that the dynamics component provides a more reliable estimate of the system behavior than the planner.
}

\changes{
In conclusion, our sampler combines the stability of operator-splitting methods with the expressiveness of diffusion-based planning and the dynamics consistency provided by the specialized dynamics model, yielding a principled balance between task fidelity and dynamic feasibility. While we do not provide a formal theoretical guarantee, our derivation offers a principled and intuitive rationale grounded in established operator-splitting theory. Developing a fully formal proof would require strong regularity assumptions on the learned score functions and transition kernels, which are difficult to verify in high-dimensional diffusion models. We therefore present this analysis as a theoretical motivation rather than a formal statement, supported by both its consistency with numerical integration theory and our empirical findings demonstrating stability and improved feasibility across diverse domains.
}

\section{Implementation Details on Unitree Go2}
\label{sec:apx-go2}
We deploy our method on a Unitree Go2 quadruped robot equipped with an onboard Jetson Orin computer, enabling fully self-contained operation without reliance on external compute resources. Executing diffusion-based planning in real time on embedded hardware is challenging due to the computational burden of the reverse sampling process. To achieve practical closed-loop control, we incorporate several system-level optimizations, similar to prior real-time diffusion-based control work~\cite{huang2025flexible}: 

\begin{itemize}
    \item \textbf{Single-sample DDIM inference:} We generate only one trajectory per planning step using DDIM, avoiding the overhead of sampling multiple candidates. 
    \item \textbf{Action chunking:} The controller executes 4 consecutive actions from the current plan before triggering replanning, amortizing the cost of trajectory generation. 
    \item \textbf{Asynchronous planning:} Diffusion sampling runs in parallel with the control loop, so future trajectories are computed in the background while the robot executes the current one. 
    \item \textbf{Warm-starting:} Instead of restarting diffusion from pure noise, we partially diffuse the previous trajectory forward for $7$ steps before denoising (see Sec.~\ref{sec:comp-budget}), reducing computation while preserving trajectory quality. 
\end{itemize}

For our diffusion models, we use a planning horizon of $H=16$ with $K=10$ denoising steps. These optimizations together enable real-time operation on the Go2, allowing control rates sufficient for agile locomotion.


\end{document}